\documentclass[letterpaper]{article}
\usepackage[preprint]{aaai2027}
\usepackage[hyphens]{url}
\usepackage{graphicx}
\usepackage{natbib}
\usepackage{caption}
\usepackage{array}
\usepackage{booktabs}
\usepackage{amsmath}
\usepackage{algorithm}
\usepackage{algorithmic}

\newcommand{\gry}[1]{#1}

\newcommand{\re}[1]{#1}

\newcommand{\apre}[1]{#1}

\newcommand{\loc}{\textsc{DCI-Loc}\xspace}
\newcommand{\qa}{\textsc{DCI-QA}\xspace}

\input{macros}

\title{\videodci{}: How Far Are VLM Agents from Human Viewers\\
in Direct Video Corpus Interaction?}
\author{
\mbox{Garry Yang\textsuperscript{1}},
\mbox{Yuqiao Liu\textsuperscript{1}},
\mbox{Hao Lin\textsuperscript{1}},
\mbox{Luohao Pan\textsuperscript{1}}\\
\mbox{Xinru Chen\textsuperscript{2}},
\mbox{Wanqi Ren\textsuperscript{2}},
\mbox{Zhikang Xu\textsuperscript{2}}\\
\mbox{Yizhan Huang\textsuperscript{1}},
\mbox{Huaixiao Tou\textsuperscript{2}},
\mbox{James Cheng\textsuperscript{1}}
}
\affiliations{
\textsuperscript{1}The Chinese University of Hong Kong\qquad
\textsuperscript{2}Independent Researcher\\
\authoremaillink{\texttt{hcyang@cse.cuhk.edu.hk}}
}

\newcommand{\authoremaillink}[1]{%
\leavevmode
\pdfstartlink attr{/Border [0 0 0] /H /N}
user{/Subtype /Link /A << /S /URI /URI (mailto:hcyang@cse.cuhk.edu.hk) >>}%
#1%
\pdfendlink
}

\makeatletter
\renewcommand{\@maketitle}{%
  \begingroup
  \centering
  \vspace*{16pt}%
  {\fontsize{16}{18}\selectfont\bfseries\@title\par}%
  \vspace{16pt}%
  {\fontsize{11.5}{14}\selectfont\bfseries\@author\par}%
  \vspace{3pt}%
  {\fontsize{11.5}{14}\selectfont\mdseries\aaai@affiliations\par}%
  \vspace{27.4pt}%
  \endgroup
}
\makeatother

\begin{document}

\maketitle
\pagestyle{plain}
\thispagestyle{plain}

\begin{abstract}
Humans answer complex visual questions by inspecting short moments, updating
what they know, and redirecting attention toward missing evidence.
Existing video benchmarks, however, often provide the relevant video, evaluate
only final answers or isolated localization, or permit textual shortcuts,
leaving multi-round visual evidence seeking underexplored.
We introduce \videodci{}, a diagnostic benchmark and functionally matched
human--agent protocol for \cmn{\textbf{D}irect Video \textbf{C}orpus
\textbf{I}nteraction}. It comprises \textsc{DCI-Loc} for temporal evidence
localization and \textsc{DCI-QA} for end-to-end discovery, retention, and
composition of cross-video evidence. In \textsc{DCI-QA}, videos within each
pool are topically related, \cm{and 75\% of the answers require hidden
temporal evidence from multiple videos}.
Given only the question and video pool, humans and agents iteratively inspect
bounded local windows, retain supporting evidence, and answer with cited
temporal spans.  Human trajectories provide reference curves for evidence
acquisition and use across rounds.
\re{Experiments reveal a consistent retrieval-to-grounding gap.  
On \loc, frontier VLMs often retrieve and expose the reference
moment, yet convert far less of this exposure into precise final citations
than the human reference.  
On \qa, the loss begins earlier and compounds across videos.
Ablation studies further confirm that iterative search outperforms
one-shot retrieval.
Multi-round search is therefore
necessary but insufficient: current agents accumulate candidates and
observations more readily than precise, jointly sufficient evidence.}
\cm{The benchmark is available at \dataurl.}

\end{abstract}

\section{Introduction}
\label{sec:introduction}

\begin{figure}[!t]
    \centering
    \includegraphics[width=\columnwidth]{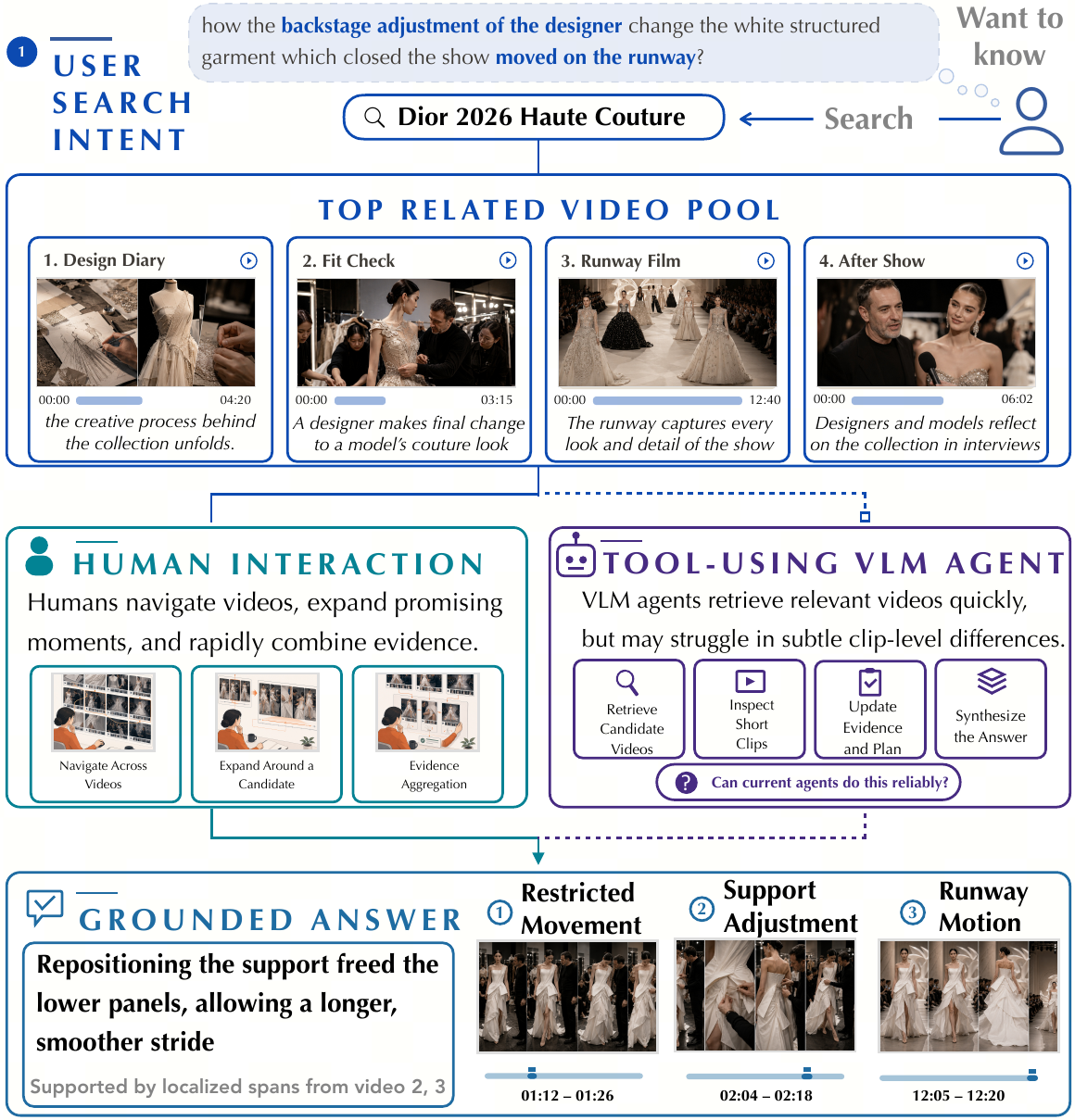}
\caption{\cmn{Direct Video Corpus Interaction}.
A search returns related videos, but each candidate is individually insufficient
for the user's question. Solvers must navigate across sources and combine
localized observations---restricted movement, the backstage support adjustment,
and freer runway motion---to ground the answer. This task asks whether tool-using VLM
agents can perform this evidence-seeking process at a level comparable to humans.}
    \label{fig:example-introduction}
\end{figure}

\cmn{Direct video corpus interaction} asks a human or agent to navigate a collection
of videos in response to a natural-language request about visual content \gry{\cite{Cunningham2008HowPeopleFindVideos, PerezOrtiz2022VideoNavigation}}.  It
involves two coupled capabilities: localizing relevant temporal evidence and,
when needed, composing that evidence into an open-ended answer grounded in
those spans \cite{Chowdhury2025MAGNET}.  
The challenge becomes more demanding when the input is a pool of topically
related videos rather than one supplied source.  
Evidence may then be
distributed across sources: one video may establish an initial state, another
show an intervention, and a third reveal the outcome.
Each observation can be relevant yet insufficient alone
\gry{\cite{Chowdhury2025MAGNET,Li2026CrossVid}}.  
Figure~\ref{fig:example-introduction} illustrates a user asking \emph{how a
backstage adjustment changed the movement of the white structured look that
closed the Dior 2026 Haute Couture show}. No single video provides the complete
evidence: a pre-adjustment clip shows restricted motion, a fitting clip shows
the support being repositioned, and a runway clip shows freer movement
afterward.
Localization requires finding an
\emph{evidence set}: a collection of mutually coherent localized spans that is
jointly sufficient for the question; grounded QA additionally requires aligning and composing them \cite{BenAmi2026HERBench}. 

Humans solve such problems through selective, iterative viewing: task goals
guide attention, local observations update an evolving interpretation, and
missing evidence redirects the next look
\cite{Rothkopf2007TaskContext,Renninger2007WhereNext,
Loschky2020SPECT,Loschky2026SPECT}.
Video agents increasingly mirror this coarse-to-fine loop: sparse observations
and retrieval indices identify candidate moments, selected regions are
inspected at higher temporal resolution, and accumulated observations guide
subsequent search
\cite{Wang2024VideoAgentLong,Jeong2025VideoRAG,
Yang2026LongVT,Shah2026VRAgent}.
With a video corpus, however, another retrieval call constitutes
progress only if it builds on prior observations---redirecting toward
complementary sources or moments, aligning observations across views, and
preserving evidence until the accumulated spans jointly support an answer \cite{Shang2024TraveLER}.
We formulate this problem in Section \ref{sec:problem}.
Repeated search becomes cumulative only when later observations resolve gaps
left by earlier ones \cite{Wang2026ActiveVideoPerception}.
To probe how VLM agents perform under video corpus interaction, we therefore ask:
\begin{quote}
\emph{Given the same question and video pool, how far are current multimodal
agents from human viewers in acquiring, retaining, and composing cross-video
evidence for a grounded answer?}
\end{quote}

Existing benchmarks do not fully support this comparison.  Standard
VideoQA primarily scores final answers; temporal-grounding benchmarks localize
relevant moments; grounded VideoQA links answers to localized support; and
recent multi-video benchmarks evaluate retrieval or reasoning over
randomly assembled video collections
\cite{Xiao2021NExTQA,Mangalam2023EgoSchema,Lei2020TVR,
Lei2021QVHighlights,Lei2020TVQAPlus,Xiao2024GroundedVideoQA,
BenAmi2026HERBench,Peng2025MVUEval,Li2026CrossVid,Feng2025MUVR,
Chowdhury2025MAGNET,Martin2026WikiVideo}.
These settings rarely combine topically related search-result pools,
open-ended questions requiring cross-video visual evidence, and round-wise
human--agent measurements of what is inspected, retained, and used.  They
therefore cannot distinguish cumulative evidence acquisition from repeated
retrieval of merely related content.

To answer this question, we introduce \videodcifull{}
(\videodci{}),\footnote{\cm{The benchmark is available at \dataurl.}} a
diagnostic benchmark suite with a matched human--agent protocol.
It comprises two complementary tasks: \re{\loc adapts single-video moment-retrieval instances from QVHighlights \cite{Lei2021QVHighlights} into ten-video candidate corpora, requiring systems
  to identify both the source video and the relevant temporal span.}
\qa evaluates
end-to-end discovery, retention, and composition of cross-video evidence for open-ended answers.  \textsc{DCI-QA} contains 80 frozen top-20
search-result pools spanning fashion, film, news, and sports, totaling \cm{1,185}
videos and 400 visual-only questions (Section \ref{sec:benchmark}).  Videos within each pool share a topical
search context and create plausible hard alternatives.  \cmn{Every question
requires a jointly sufficient set of temporally localized facts, and for 299 of
the 400 that set spans at least two videos}.  Solvers see only the question and video pool; construction seeds remain hidden.
Independent review
verifies joint sufficiency \cm{and, for the cross-video items, multi-video
necessity}.

\re{During evaluation, humans and agents follow a functionally matched
visual-only protocol.  In each macro-round, they select a source and temporal
location, inspect a bounded local window,
update retained evidence, and either search again or answer.  We evaluate
answer quality and complete evidence localization separately, while tracing candidate retrieval,
evidence exposure, retention, and final citation across rounds.
Results reveal a consistent retrieval-to-grounding gap.  
In
\loc, agents usually find the target video and often expose a
reference moment, but many such observations never become correct,
temporally precise citations; humans are substantially more effective at
converting exposure into final localization.  On the
\qa task, agents retrieve plausible sources
comparable to the human reference, yet expose and cite fewer answer-bearing
spans and produce weaker answers.  Ablation study shows that iterative search substantially improves
coverage over one-shot retrieval and efficiency over uniform skimming.
Oracle access shows that both source
discovery and temporal localization remain important, but the weak answer quality under \textsc{Oracle-Span} \cmn{suggests that visual interpretation and
cross-video synthesis remain bottlenecks}.  We conclude that current VLM agents accumulate observations more reliably than precise,
complete, answer-supporting evidence.}
Our contributions are threefold:
\begin{itemize}
    \item We formulate \cmn{Direct Video Corpus Interaction} and introduce
    \videodci{}, pairing \loc for temporal evidence
    localization with \qa for cross-video evidence discovery, composition, and grounded answering.

    \item We construct \qa from 80 frozen top-20 search-result
    pools, comprising \cm{1,185} videos and 400 visual-only questions with localized
    evidence, \cm{verified multi-video necessity for the 299 cross-video
    items}, and realistic hard alternatives.

    \item We define a matched human--agent protocol and stage-wise
    diagnostics spanning candidate retrieval, evidence exposure, retention,
    final citation, and answer quality, revealing that the main gap lies in
    converting plausible observations into precise, jointly sufficient
    evidence.
\end{itemize}

\section{Related Work}
\label{sec:related-work}

\paragraph{Video Agents.}
Current video agents commonly follow a coarse-to-fine, multi-round process:
they retrieve candidate videos or moments from sparse observations or an
index, inspect promising regions at higher temporal resolution, and search
again when the accumulated evidence remains incomplete
\cite{Wang2024VideoAgentLong,Shang2024TraveLER,Jeong2025VideoRAG,
Yang2026LongVT,Wang2026ActiveVideoPerception}.
Candidate discovery may rely on embedding similarity, predicted timestamps,
or query-specific evidence decomposition
\cite{Wang2024VideoAgentLong,Shang2024TraveLER,
Yang2026LongVT,Yan2026Symphony}.
Inspected clips are then summarized into an evidence state
that conditions subsequent observations and the final answer.  This process
resembles the observe--update--act loop used by general LLM agents
\cite{Yao2023ReAct,Zou2026BeliefDeviation}.  However, existing video-agent evaluations primarily
report final answer or retrieval performance.
\re{They rarely measure whether useful visual evidence is discovered and retained
across rounds, or how efficiently this process compares with human viewers.}

\paragraph{Evidence-Grounded Benchmarks.}
Standard VideoQA benchmarks evaluate causal, temporal, or long-form reasoning
\cite{Xiao2021NExTQA,Mangalam2023EgoSchema}.  Answer accuracy alone, however,
does not expose whether the prediction is supported by visual evidence
\cite{mesh,Xiao2024GroundedVideoQA}.
Temporal-grounding benchmarks localize relevant moments, while grounded
VideoQA associates answers with spatial or temporal evidence
\cite{Lei2020TVR,Lei2021QVHighlights,Lei2020TVQAPlus,
Xiao2024GroundedVideoQA,BenAmi2026HERBench}.
Most benchmarks nevertheless focus on a supplied single video.
Recent benchmarks extend video understanding to preassembled or curated
collections, including MVU-Eval, CrossVid, MUVR, AVHaystacks, and WikiVideo
\cite{Peng2025MVUEval,Li2026CrossVid,Feng2025MUVR,
Chowdhury2025MAGNET,Martin2026WikiVideo}.
\re{They leave underexplored a setting in which related search results contain
plausible but insufficient alternatives, and a solver must iteratively locate
and combine complementary visual evidence.  \videodci{} targets this setting
with search-result pools and open-ended visual-only questions.  It
jointly evaluates answer correctness, complete evidence localization, and
round-wise evidence accumulation, including a matched human reference study.}

\section{Problem Formulation}
\label{sec:problem}

\subsection{Evidence Video Corpus Interaction}

Let $q$ be an open-ended visual question and
\begin{equation}
\mathcal{V}=\{v_1,\ldots,v_n\}
\end{equation}
a query-specific corpus of topically related videos.  
We use
\emph{solver} to refer to either a human participant or a VLM agent.  The
solver receives only
\begin{equation}
x=(q,\mathcal{V}),
\end{equation}
without being told which videos contain relevant evidence or where the
relevant moments occur.

A temporal evidence span is a source-video locator
\begin{equation}
p=\bigl(v,[t^{\mathrm{s}},t^{\mathrm{e}}]\bigr),
\qquad v\in\mathcal{V},
\end{equation}
and an evidence set $P=\{p_1,\ldots,p_r\}$ contains the spans cited for an
answer.  We write
\begin{equation}
\operatorname{Suff}(q,a,P)=1
\end{equation}
when the visible facts established by $P$ jointly support every material
component of answer $a$.  

The interaction yields
\begin{equation}
\widehat{y}=
\bigl(\widehat{a},P,\tau\bigr),
\end{equation}
where $\widehat{a}$ is the answer to question $q$, $P$ is its cited visual evidence,
and $\tau$ is the externally visible search trajectory.  A fully successful
prediction must therefore provide both a correct answer and a sufficient
evidence set.

We call an evaluation unit \emph{strictly multi-video} when the corpus contains
sufficient evidence, but no evidence drawn from any individual video supports
the complete answer:
\begin{equation}
\begin{split}
&\exists P\subseteq\operatorname{Spans}(\mathcal{V}) :
\operatorname{Suff}(q,a^*,P)=1,\\
&\forall v_i\in\mathcal{V},\
\forall P_i\subseteq\operatorname{Spans}(v_i):
\operatorname{Suff}(q,a^*,P_i)=0.
\end{split}
\label{eq:strict-multivideo}
\end{equation}

\subsection{Viewer-Inspired Multi-Round Protocol}

Human viewing provides a functional reference for this interaction.  Given a
task, viewers need not process every video sequentially.  They can selectively
sample candidate sources and moments, update their current interpretation from
local observations, and redirect attention when evidence remains missing or
inconsistent
\cite{Loschky2020SPECT,Loschky2026SPECT}.  We adopt this
observation--state--attention loop as the basis of our protocol.

Accordingly, both a human participant and a video agent solve an item through
a sequence of macro-rounds.  In each round, the solver selects a video and
temporal location, inspects a local clip, updates its evidence state, and then
either searches again or answers.

At round $t$, the agent exposes a public state
\begin{equation}
m_t=
\bigl(\mathcal{L}_t,P_t,\widetilde{a}_t,U_t\bigr),
\label{eq:public-agent-state}
\end{equation}
where $\mathcal{L}_t$ records all inspected spans, $P_t\subseteq\mathcal{L}_t$
contains the evidence retained for answering, $\widetilde{a}_t$ is the current
answer hypothesis, and $U_t$ contains the unresolved evidence requirements.
Initially,
\begin{equation}
\mathcal{L}_0=P_0=\emptyset,
\quad
\widetilde{a}_0=\bot,
\quad
U_0=\operatorname{Decompose}(q).
\end{equation}

At each round, the agent performs
\begin{equation}
\begin{array}{l}
u_t=\operatorname{Plan}(q,m_{t-1};b_t),\\[2pt]
C_t=\operatorname{Search}(u_t,\mathcal{V}),\\[2pt]
O_t=\operatorname{Inspect}(q,C_t,m_{t-1}),\\[2pt]
m_t=\operatorname{Update}(m_{t-1},O_t),
\end{array}
\label{eq:agent-loop}
\end{equation}
where $b_t$ is the remaining budget, $u_t$ specifies the next source or
temporal region to inspect, $C_t$ contains retrieval candidates, and $O_t$
contains observations from the selected raw video clips.

The protocol distinguishes retrieval from evidence.  A candidate in $C_t$ may
be topically relevant but temporally inaccurate or visually insufficient.  It
enters $P_t$ only after local inspection establishes a concrete visual fact.
Likewise, an isolated frame cannot establish unseen motion, continuity, or a
before--after relation.

The trajectory is
\begin{equation}
\tau=\{(u_t,C_t,O_t,m_t)\}_{t=1}^{T}.
\end{equation}
A grounded termination occurs when
\begin{equation}
\operatorname{Suff}
\bigl(q,\widetilde{a}_T,P_T\bigr)=1
\qquad\text{and}\qquad
U_T=\emptyset.
\end{equation}
Both humans and agents may instead reach a search stall: for example, they may
revisit related but insufficient clips, locate the correct video but miss the
decisive interval, or retain only part of the required evidence.  With
remaining budget, a well-behaved solver should redirect attention toward the
unresolved requirement by switching sources, expanding a temporal window, or
seeking another viewpoint. \re{All evaluation units are assumed to be answerable within the given video corpus.}

This formulation separates three capabilities:
\emph{evidence acquisition}, whether relevant spans enter $\mathcal{L}_t$;
\emph{evidence incorporation}, whether inspected evidence is retained in
$P_t$ and changes the public state; and \emph{evidence composition}, whether
the retained spans jointly support the final answer.  
\re{Section~\ref{sec:benchmark}
describes the supervision used to evaluate these stages.}

\begin{figure*}[t]
    \centering
    \includegraphics[width=\textwidth]{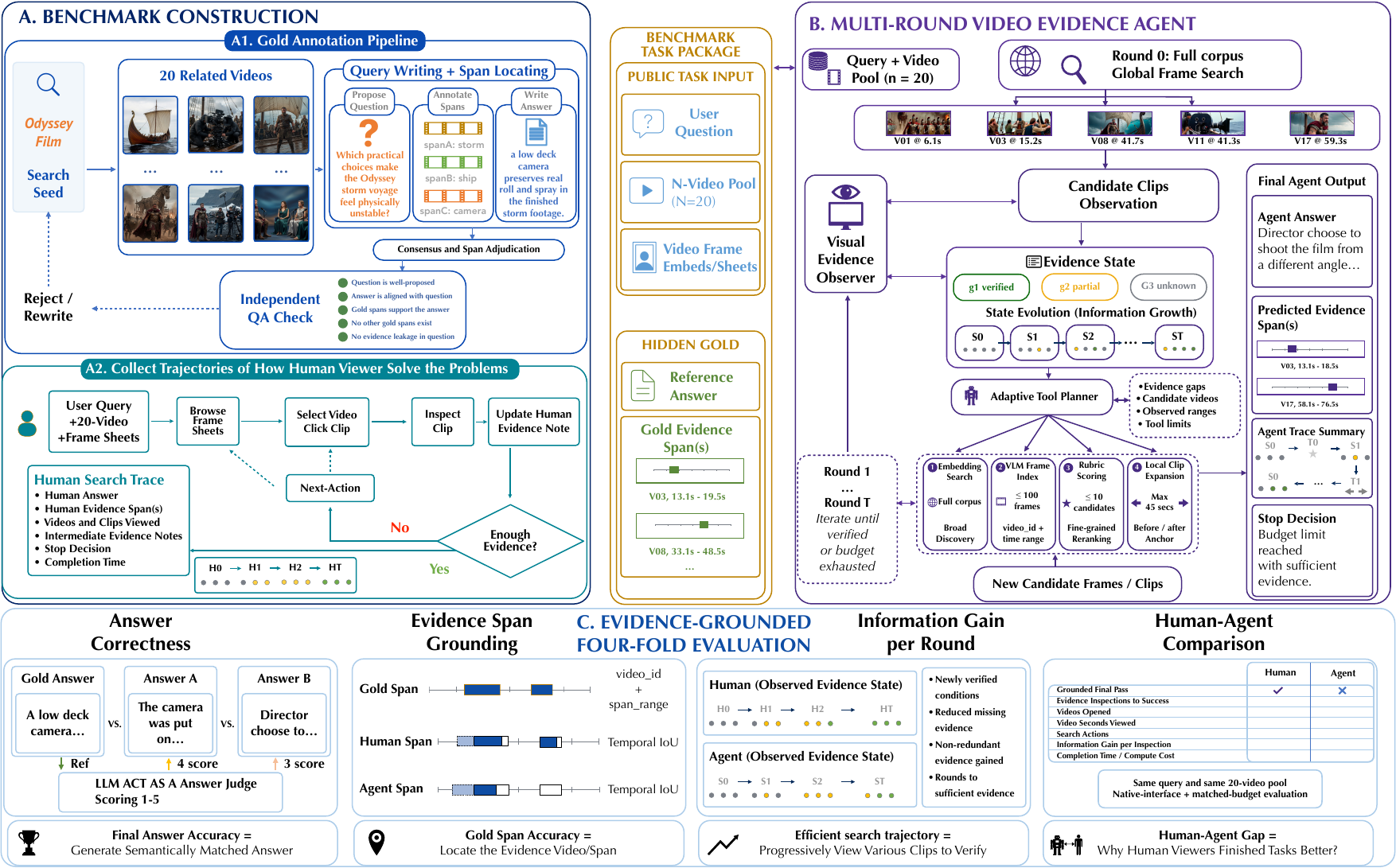}
    \caption{Overview of Video-DCI-QA construction, interaction, and evaluation.
    (A) Annotators inspect a 20-video search-result pool, author a
    question requiring complementary cross-video evidence, localize
    answer-supporting spans, and independently verify question validity and
    evidence sufficiency and necessity; the corpus interface records search
    trajectories of another human evaluator.  (B) Given only the question and video pool,
    an agent iteratively retrieves candidates, inspects clips, updates its
    evidence state, and returns an answer with cited spans, while reference
    answers and spans remain hidden.  (C) Agent answers are
    compared with ground-truth and human performance by correctness, temporal grounding, evidence accumulation,
    and efficiency.}
    \label{fig:example-benchmark}
\end{figure*}

\section{Benchmark Design}
\label{sec:benchmark}

\videodcifull{} (\videodci{}) evaluates direct interaction with a corpus of
related videos.  Following Section~\ref{sec:problem}, the solver receives only
a visual query and the corresponding video corpus.  Construction methods,
reference video identities, temporal locations, evidence roles, and answers
remain hidden.

\subsection{Construction}

\paragraph{Two complementary tracks.}
We construct \textbf{\textsc{DCI-Loc}} from 500 QVHighlights validation queries
\citep{Lei2021QVHighlights}. Each query is placed in a fixed ten-video
candidate corpus containing its source video, and the solver must retrieve the
source video and localize the relevant temporal moment.
\textbf{\textsc{DCI-QA}} is the main evidence-grounded question-answering track.  We
construct 80 corpus groups from YouTube search seeds spanning fashion, film,
news, and sports-event footage.  For each seed, we freeze a twenty-video pool
after pre-specified technical-quality and near-duplicate screening, while
preserving the original ranks and exclusion decisions.

\re{After reviewing each corpus, annotators write five open-ended,
visual-only questions about events, states, transformations, or relations
distributed across its videos.}  Questions may describe semantic stages of an
event, but do not disclose gold video identities.
Videos within
a pool are topically related and therefore provide plausible search
alternatives.
\cm{\textsc{DCI-QA} contains 80 corpora, 1,185 videos, and 400 questions.
On average, each question pool contains 14.8 videos with a total available
duration of 39 minutes.}
Full construction details and statistics are reported in Appendix~\ref{app:benchmark-details}.

\paragraph{Evidence supervision and verification.}
Each QA question is annotated with a reference answer, semantic evidence
roles, and a primary set of temporally localized spans.  A role specifies a
visual fact required by the answer; each span records its source video,
temporal interval, and the fact visible within that interval.  \cm{The benchmark
contains 1,201 reference spans, three per question, with a mean duration of
23.1 seconds.}
A second annotator verifies that the primary set is jointly sufficient and
necessary, and searches the full corpus to eliminate alternative support.  \cm{We remove
questions that require the audio track and questions that frame-level OCR
alone can answer}
(Appendix~\ref{app:necessity-verification}).

\subsection{Evaluation Protocol}
\label{sec:benchmark-evaluation}

\paragraph{Observable evidence stages.}
Across both tracks, we distinguish a retrieved candidate, the frames or clip
actually inspected, supporting evidence retained from that observation, and
the spans submitted as final citations.  Only visual content exposed to the
solver counts as inspected evidence.
These
stages separate candidate discovery, evidence exposure, evidence
incorporation, and final grounding.

\paragraph{Track-specific outcome measures.}
\textsc{DCI-Loc} reports whether the reference video is reached at the
candidate, displayed-frame, retained-state, and final-citation stages, together
with mean temporal IoU and success at IoU~0.5.  Because \textsc{DCI-QA}
requires several evidence members, the corresponding source and span measures
are recalls over the reference set: candidate-video recall, displayed-span
recall, retained-video recall, and cited-video recall.  Final span coverage is
computed by one-to-one same-video matching, so that one prediction cannot
satisfy several reference spans.  Two blinded LLM judges assign each submitted
answer an anchored score from 1 to 5; their mean is reported as Ans.-5.

\paragraph{Round-wise evidence use.}
At each macro-round cutoff, we recompute how much reference evidence has been
exposed by the displayed media.  \re{We additionally trace when evidence that
appears in the final citations first became available, yielding cumulative
seen and ultimately cited curves.}  Their difference measures the failure to
convert encountered evidence into final grounded output.  We report these
curves over macro-rounds, together with total rounds and active latency.
Answer quality is judged based on the final response.

\paragraph{Human reference study.}
Three participants complete 50 questions per track under the same visual-only
corpus setting.  Titles, descriptions, transcripts, OCR indices, and generated
captions are hidden.  Participants navigate a low-rate storyboard, select a
video and temporal location in each macro-round, inspect a local clip, and
either inspect another location, expand the current region, or submit an
answer with cited evidence.  Human performance is reported as a reference
under the comparable source, localization, answer, and round-wise seen--cited
measures.  \re{Full procedures are given in
Appendix~\ref{app:human-protocol}.}

\section{Experiments and Analysis}
\label{sec:experiments}

We organize the experiments around three questions:
(1) \cmn{how well do current VLM agents perform in direct video corpus interaction};
(2) how do humans and agents acquire and use evidence across rounds; and
(3) how do adaptive search and evidence access affect agent
performance?

\subsection{Experimental Setup}

\paragraph{Systems.}

We compare three visual-only strategies under matched corpus access.
We implement \textsc{Iterative Agent} which repeatedly
searches, observes, updates its evidence state, and replans.
We evaluate 10 models using the same cached visual
index, observation interface, and hard budgets.
The main track disables web search, titles, descriptions, transcripts, OCR
indices, and generated captions.
We additionally report two diagnostic oracles.
\textsc{Oracle-Video} reveals the videos in one acceptable evidence set but
not their timestamps, whereas \textsc{Oracle-Span} supplies the reference
windows directly.
The two oracle gaps diagnose source selection, temporal localization, and
residual visual interpretation or cross-video composition errors.

\paragraph{Iterative evidence agent.}
The agent first converts the question into an immutable contract of answer
components $\mathcal{K}$.
Round~0 uses role-conditioned \texttt{initial\_observer} (embedding similarity/corpus skim) to retrieve initial
candidates $u$.  
After observation, the agent updates
the evidence state $M$, and only then invokes
the planner. The planner selects one of the action set $\mathcal{A}$ that normalize common coarse-to-fine video-agent primitives.
\texttt{atomic\_embedding} searches the full corpus when the missing evidence
can be expressed as a concise visual phrase; \texttt{global\_skim} directly
searches 100 cached frames across the corpus;
\texttt{vlm\_region\_index} localizes evidence inside a known video range;
\texttt{rubric\_candidate\_rerank} discriminates among 1--10 already localized
candidates; and \texttt{expand\_current\_observation} enlarges a promising
window when the missing endpoint or result is likely nearby.  
The agent actively updates its retained evidence and unresolved needs, and a
set verifier checks answer-contract support.  It continues while evidence remains missing and budget remaining.  Once the
retained set is sufficient, the controller generates the answer and maps its
supporting evidence to cited temporal spans; if the budget is exhausted or no
legal action remains, it returns a calibrated incomplete output.  \re{Implementation details are given in
Appendix~\ref{app:experiment}.}

\begin{table*}[t]
\centering
\caption{Stage-wise \textsc{DCI-Loc} and \textsc{DCI-QA}
performance.
\textsc{DCI-Loc} uses three QVHighlights repetitions.
\textsc{DCI-QA} reports the outcome-conditioned Full Corpus diagnostic over two repetitions;
Gemma 4 31B evidence/efficiency uses three repetitions of the 115-case appendix subset.
Ans.-5 averages GPT-5.4/GPT-5.6 scores (1--5).
Rounds and Time are mean macro-rounds and median active latency.
Human references use 50 tasks per track.
}
\label{tab:dci-main-combined}
{\small
\setlength{\tabcolsep}{0.8mm}
\begin{tabular*}{\textwidth}{@{\extracolsep{\fill}}lccccccccccc@{}}
\toprule
\multicolumn{1}{c}{\shortstack{Metric\\$\uparrow$}}
& \multicolumn{1}{c}{\shortstack{GPT\\4o}}
& \multicolumn{1}{c}{\shortstack{GPT\\5.2}}
& \multicolumn{1}{c}{\shortstack{GPT\\5.4}}
& \multicolumn{1}{c}{\shortstack{GPT\\5.5}}
& \multicolumn{1}{c}{\shortstack{GPT\\5.6 Sol}}
& \multicolumn{1}{c}{\shortstack{Gemini\\2.5 Pro}}
& \multicolumn{1}{c}{\shortstack{Gemini\\3 Pro}}
& \multicolumn{1}{c}{\shortstack{Gemini\\3.5 Flash}}
& \multicolumn{1}{c}{\shortstack{Gemma 4\\12B}}
& \multicolumn{1}{c}{\shortstack{Gemma 4\\31B}}
& \multicolumn{1}{c}{\shortstack{Human\\Reference}} \\
\midrule
\multicolumn{12}{@{}l}{\textbf{\textsc{DCI-Loc}}: Temporal Localization} \\
\cmidrule(lr){1-3}
CandV-Hit & 83.5 & 88.4 & 87.7 & 89.0 & \textbf{91.1} & 84.6 & 84.7 & \underline{90.6} & 75.3 & 72.7 & 88.7 \\
FrameHit & 70.2 & 79.3 & 77.3 & 81.5 & \textbf{84.7} & 72.4 & 74.0 & \underline{83.5} & 66.6 & 66.1 & 79.6 \\
CiteV-Acc & 20.9 & 38.1 & 49.5 & \underline{59.2} & 57.1 & 58.7 & 58.1 & \textbf{71.2} & 33.9 & 50.9 & \textbf{85.9} \\
mTIoU & 9.10 & 19.8 & 23.9 & \underline{25.8} & 25.7 & 24.1 & 23.8 & \textbf{29.2} & 11.9 & 19.7 & \textbf{51.9} \\
TIoU@.5 & 8.50 & 20.7 & 24.7 & \underline{26.5} & 26.1 & 24.5 & 24.5 & \textbf{28.0} & 11.8 & 20.0 & \textbf{55.3} \\
Rounds $\downarrow$ & \textbf{3.91} & 6.18 & \underline{4.48} & 6.31 & 5.62 & 4.99 & 5.13 & 5.32 & 5.91 & 5.16 & 2.14 \\
Time (min) $\downarrow$ & \textbf{1.32} & 2.33 & 1.59 & 3.28 & 1.99 & 4.09 & 4.10 & 2.56 & \underline{1.56} & 2.43 & 1.74 \\
\midrule
\multicolumn{12}{@{}l}{\textbf{\textsc{DCI-QA}}: Cross-Video Evidence QA} \\
\cmidrule(lr){1-3}
CandV-Rec & 46.7 & 63.0 & 69.6 & \textbf{71.3} & \underline{70.3} & 66.1 & 70.1 & 60.3 & 51.7 & 62.8 & 61.1 \\
SpanHit-Rec & 7.80 & 38.3 & \underline{39.0} & 38.0 & \textbf{40.4} & 32.2 & 32.2 & 32.4 & 17.2 & 28.6 & \textbf{52.4} \\
RetV-Rec & 14.1 & 54.9 & 59.7 & \underline{60.4} & \textbf{63.2} & 55.1 & 51.7 &  46.1 & 30.4 & 48.3 & -- \\
CiteV-Rec & 12.6 & 39.7 & 42.9 & 44.8 & \textbf{46.8} & 41.6 & 37.7 & \underline{46.1} & 27.1 & 44.3 & \textbf{58.6} \\
SpanCov & 2.90 & 8.40 & 8.20 & \underline{10.0} & \textbf{10.6} & 8.10 & 8.50 & 9.10 & 4.90 & 8.21 & \textbf{23.8} \\
Ans.-5 & 1.20 & 1.91 & 2.31 & \textbf{2.61} & \underline{2.53} & 2.03 & 1.87 & 2.25 & 1.50 & 1.81 & \textbf{3.46} \\
Rounds $\downarrow$ & \textbf{2.28} & 10.8 & 9.78 & 9.24 & 10.2 & 8.66 & 8.50 & 10.2 & \underline{3.31} & 6.16 & 8.50 \\
Time (min) $\downarrow$ & \textbf{2.59} & 6.98 & \underline{3.90} & 5.53 & 4.55 & 10.2 & 8.08 & 6.81 & 4.74 & 8.02 & 10.6 \\
\bottomrule
\end{tabular*}}

\end{table*}

\begin{algorithm}[t]
\small
\caption{Iterative visual evidence search.}
\label{alg:evidence-search}
\begin{algorithmic}[1]
\REQUIRE question $q$, video corpus $\mathcal{V}$, budget $B$
\STATE $K \leftarrow \operatorname{Contract}(q)$;
       $M \leftarrow \emptyset$
\STATE $\begin{aligned}[t]
\mathcal{A}\leftarrow\{&
\operatorname{AtomicEmbedding},
\operatorname{GlobalSkim},\\
&\operatorname{RegionIndex},
\operatorname{CandidateRerank}, \operatorname{Expand}
\}
\end{aligned}$
\STATE $u \leftarrow
       \operatorname{InitialAction}(K,\mathcal{V})$
       \textit{(Round 0)}
\WHILE{$B>0$}
\STATE $(O,B) \leftarrow \operatorname{Observe}(u,B)$
\STATE $M \leftarrow \operatorname{Update}(M,O)$
    \IF{$\operatorname{Sufficient}(K,M)$}
        \STATE \textbf{return}
        $\operatorname{AnswerAndCite}(q,M)$
    \ELSIF{$B\leq 0 \lor \operatorname{NoLegalAction}(M)$}
        \STATE \textbf{return}
        $\operatorname{CalibratedIncomplete}(q,M)$
    \ELSE
        \STATE $u \leftarrow \operatorname{Planner}(M,\mathcal{A})$
    \ENDIF
\ENDWHILE
\STATE \textbf{return}
$\operatorname{CalibratedIncomplete}(q,M)$
\end{algorithmic}
\end{algorithm}

\paragraph{Evaluation details.}
\re{\textsc{DCI-Loc} evaluates QVHighlights validation queries in fixed
ten-video cohorts over three repetitions, with at most 16 macro-rounds.
\textsc{DCI-QA} reports the outcome-conditioned diagnostic
under Full Corpus access.
Within each track, models share the same visual-only access, cached index,
observation interface, and hard budgets.
We report stage-wise source or span retrieval, exposure, retention, and
citation, together with temporal localization or coverage, Ans.-5, rounds,
and active latency.
Ans.-5 averages two LLM judges.
Human--agent trajectories compare cumulative seen and ultimately cited
evidence.
Details are given in Appendix~\ref{app:evaluation-details}.}

\subsection{Results and Analysis}

\paragraph{RQ1: \cmn{How well do current VLM agents perform in direct video corpus interaction?}}

Table~\ref{tab:dci-main-combined} reports the overall evidence-grounding and VQA accuracy on \videodci.
In \loc, VLM agents generally retrieve the target video (CandV-Hit) and often
expose a frame within a reference moment (FrameHit), but this early success does not
reliably propagate to the final citation (CiteV-Acc).  
Correct-video citation and precise temporal overlap
drop substantially, indicating that \textbf{discovering a plausible source is easier
than identifying and preserving the interval in answers}.  
GPT-5.6 performs
best at candidate discovery and frame exposure, while Gemini 3.5 Flash achieves the
strongest final localization.  
GPT-5.4 approaches the
best final localization with fewer rounds and lower latency, whereas GPT-4o's
short trajectories coincide with substantially weaker grounding.
Among the open-weight
models, scaling Gemma from 12B to 31B mainly improves downstream citation
quality.

This gap between exposure and citation becomes more pronounced in \qa, where the solver
must assemble complementary evidence across videos. 
Strong agents retrieve
many of the required sources and retain a substantial portion of them (CandV-Rec, SpanHit-Rec), but
they expose fewer reference moments and cover only a small fraction of the
required temporal spans in their retention and final citations (RetV-Rec, CiteV-Rec, SpanCov).  
GPT-5.5 and GPT-5.6 are the
strongest overall: GPT-5.6 leads most evidence-acquisition and
citation measures, while GPT-5.5 obtains the highest answer score.  
The
remaining mismatch between source recall, span coverage, and answer quality
suggests that \textbf{agents often answer from partially relevant source-level
evidence without reconstructing a temporally complete evidence set}.  
The human
reference does not dominate candidate discovery, but is
markedly stronger on downstream localization and answer quality.
This places the main human--agent gap after retrieval.

\begin{figure}[!t]
    \centering
    \includegraphics[width=\columnwidth]{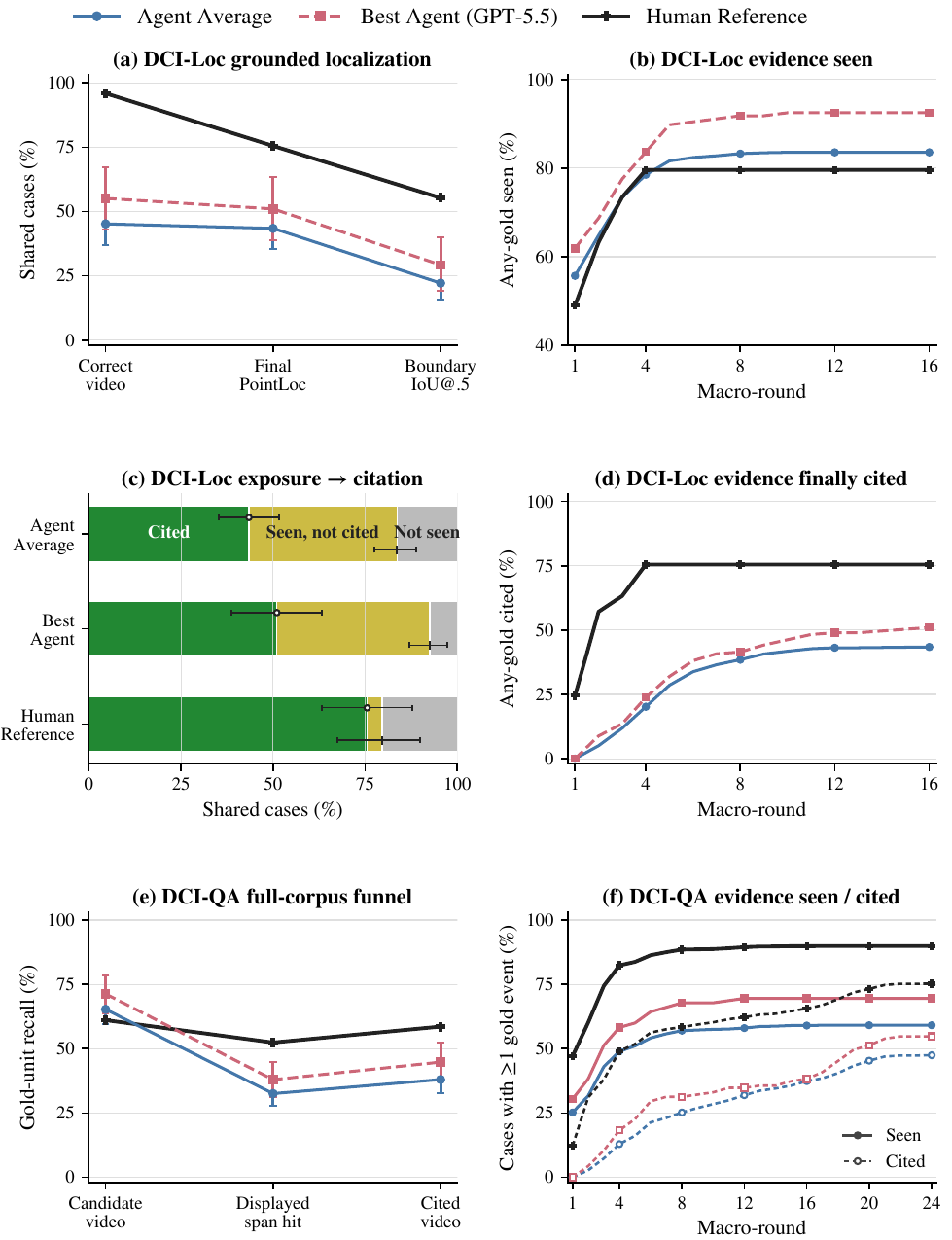}
    \par\smallskip
    \caption{\re{Human–Agent Comparison of Multi-Round Evidence Discovery, Temporal Localization, and Grounded Citation across \textsc{DCI-Loc} and Cross-Video \textsc{DCI-QA}.}}
    \label{fig:gap-decomposition}
\end{figure}

\paragraph{RQ2: How do humans and agents acquire and use evidence across rounds?}
Figure~\ref{fig:gap-decomposition} compares how reference evidence is exposed
and converted into final citations over macro-rounds.
On \loc, agents expose reference evidence quickly and eventually
see as much as, or more than, the human reference.  Their citation curves,
however, rise much more slowly: a large share of exposed evidence is never
used in the final localization.  Humans show the opposite pattern.  Their seen
coverage plateaus after only a few rounds, but nearly every exposed reference
moment is converted into a final citation, yielding much stronger source
selection and temporal overlap.  
The localization gap therefore arises 
mostly from failing to recognize and preserve evidence after exposure.
On \qa, the gap appears earlier and compounds
across the evidence funnel.  Agents retrieve plausible candidate videos at
rates comparable to the human reference, yet expose fewer answer-bearing spans
and cite fewer required sources.  Human seen and cited coverage grows rapidly
over the first several rounds, whereas agent gains become progressively
smaller and saturate at a lower level.  
Across both tasks, current agents are
therefore better at identifying plausible sources than at turning them into
localized, answer-supporting evidence: \loc primarily reveals an
evidence-use bottleneck, while \qa adds a substantial
evidence-acquisition bottleneck.

\paragraph{RQ3: How do adaptive search and evidence access affect agent performance?}
\begin{table}[t]
\centering
\caption{\textsc{DCI-Loc} search-policy and evidence-access
diagnostics (Gemini 3 Pro).
\textsc{Uniform-Skim}, \textsc{One-Shot}, and \textsc{Iterative Agent} compare
fixed, single-step, and adaptive search; the oracle conditions reveal the
reference video or span only to the controller.}
\label{tab:dci-loc-policy-ablation}
{\footnotesize
\setlength{\tabcolsep}{2.2pt}
\begin{tabular}{@{}lrrrrr@{}}
\toprule
Policy
& FrameHit 
& FinalV 
& R-Cov. 
& IoU@.5 
& Rnds. $\downarrow$ \\
\midrule
Uniform-Skim & 94.0 & 46.8 & 37.4 & 29.2 & 10.54 \\
One-Shot & 49.0 & 20.6 & 9.9 & 10.0 & 1.45 \\
\textbf{Agent} & 74.0 & 56.0 & 24.4 & 23.0 & 5.05 \\
\midrule
Oracle-Video & 90.4 & 73.2 & 32.0 & 29.0 & 5.01 \\
Oracle-Span & 99.0 & 69.0 & 47.8 & 48.2 & 1.96 \\
\bottomrule
\end{tabular}}

\end{table}

\begin{table}[t]
\centering
\caption{\textsc{DCI-QA} evidence-access diagnostics (GPT-5.5).
\textsc{Oracle-Video} supplies the reference videos, while
\textsc{Oracle-Span} additionally supplies their temporal windows; all
downstream observation, retention, citation, and answer stages remain active.}
\label{tab:dci-qa-evidence-access}
{\footnotesize
\setlength{\tabcolsep}{1.8pt}
\begin{tabular*}{\columnwidth}{@{\extracolsep{\fill}}lrrrrrr@{}}
\toprule
Policy
& \shortstack{SpanHit\\Rec. $\uparrow$}
& \shortstack{CiteV\\Rec. $\uparrow$}
& \shortstack{Span\\Cov. $\uparrow$}
& \shortstack{Span\\@.5 $\uparrow$}
& \shortstack{Ans.\\5 $\uparrow$}
& \shortstack{Rnds.\\$\downarrow$} \\
\midrule
\textbf{Full} & 38.0 & 44.8 & 10.0 & 7.8 & 2.61 & 9.24 \\
Oracle-Video & 57.0 & 77.2 & 18.3 & 14.5 & 2.85 & 5.63 \\
Oracle-Span & 85.2 & 78.3 & 67.6 & 65.5 & 3.00 & 3.38 \\
\bottomrule
\end{tabular*}}

\end{table}

On \loc in Table~\ref{tab:dci-loc-policy-ablation}, the iterative agent substantially outperforms
\textsc{One-Shot}, confirming that later evidence-search rounds are necessary.
\textsc{Uniform-Skim} exposes reference moments more often through dense
coverage, but is less efficient and less reliable in selecting the final
source.  Thus, seeing a relevant frame does not guarantee that the system
recognizes, retains, and cites the correct evidence.  The adaptive agent
achieves a better source-selection trade-off.
The oracle results separate the remaining bottlenecks.  
Revealing the correct
video improves both localization and efficiency, showing that source discovery
remains difficult.  
Providing the reference span yields much larger temporal
coverage gains, especially on \qa in Table~\ref{tab:dci-qa-evidence-access},
but produces only a modest
increase in answer quality.  Evidence access therefore explains much of the
grounding error, while residual failures remain in visual interpretation and cross-video answer synthesis.

\section{Limitations and Future Work}
\label{sec:limitations}

Although source videos in \qa are
limited to 2026 uploads,
updated models may have encountered the original video contents; after release, benchmark items may also enter later training corpora.
Our iterative agent instantiates common coarse-to-fine tool families;
better query decomposition, evidence memory, tool scheduling, or stopping
policies may further narrow the gap.  
Human evidence search is stochastic and interface-dependent: familiarity,
attention, and strategy can alter search trajectories.
Future work may develop long-horizon
VLM agents that 
learn cumulative
search policies from trajectory-level feedback.

\section{Conclusion}
\label{sec:conclusion}
\re{\cmn{Direct Video Corpus Interaction} requires more than retrieving a plausible
source: a solver must expose the answer-bearing moments, preserve and combine
compatible evidence, and ground its answer in cited temporal spans.  We
introduce \videodci{}, pairing \textsc{DCI-Loc} with \textsc{DCI-QA} and a
functionally matched human--agent protocol to evaluate these stages under
visual-only access.  Experiments reveal a persistent
retrieval-to-grounding gap.  On \textsc{DCI-Loc}, frontier agents generally
retrieve the target video and often encounter a reference moment, yet convert
far less of this exposure into accurate temporal citations than the human
reference.  On \qa, the loss compounds across evidence acquisition, citation, and answer
generation.  
Oracle results further expose remaining bottlenecks in source
discovery, temporal localization, and evidence composition.
Multi-round search is therefore necessary but insufficient: iterative search
improves coverage over one-shot retrieval, yet even oracle reference spans
yield only modest gains in answer quality.  By tracing retrieval, exposure,
retention, and citation across rounds, \videodci{} reveals where evidence is
lost rather than only whether the answer is correct, while matched human
trajectories provide reference curves for evidence acquisition and use.
Future VLM agents should learn to recognize and preserve answer-bearing
moments, turning observations into precise, jointly sufficient evidence.
\cm{The benchmark is available on Hugging Face at \dataurl.}
}

\bibliography{references}

\clearpage
\appendix
\setcounter{secnumdepth}{2}
\section{Extended Problem Formulation}
\label{app:extended-problem}

This appendix clarifies the supervision and evidence semantics underlying
Section~\ref{sec:problem}.

\subsection{Reference Supervision and Evidence Sets}

For each question $q$, the benchmark stores a hidden reference answer $a^*$,
semantic evidence roles, and localized reference spans:
\begin{equation}
\begin{gathered}
\mathcal{E}^{*}(q)=\{e_1,\ldots,e_m\},
\qquad
\mathcal{C}^{*}(q)=\{c_1^{*},\ldots,c_K^{*}\},
\\
c_j^{*}=\bigl(v_{i(j)},[t_j^{\mathrm{s}},t_j^{\mathrm{e}}]\bigr).
\end{gathered}
\label{eq:appendix-reference-structure}
\end{equation}
Each role specifies a visual fact needed by the answer, and each span is paired
with a concise statement of the fact visible within that interval.  Their
support relation is generally many-to-many:
\begin{equation}
\Gamma^{*}(q)
\subseteq
\mathcal{E}^{*}(q)\times\mathcal{C}^{*}(q).
\label{eq:appendix-role-span-relation}
\end{equation}
Thus, one span may support several roles, while one role may require several
observations or viewpoints.

The annotation may contain one or more acceptable evidence sets,
\begin{equation}
    \begin{gathered}
\mathcal{S}^{*}(q)=\{S_1^{*},\ldots,S_L^{*}\},
\qquad
S_\ell^{*}\subseteq\mathcal{C}^{*}(q),
\qquad\\
\operatorname{Compat}(q,S_\ell^{*})
=
\operatorname{Suff}(q,a^*,S_\ell^{*})
=1.
\end{gathered}
\label{eq:appendix-acceptable-sets}
\end{equation}
Here $\operatorname{Compat}$ requires the selected spans to admit the
composition requested by the question: for example, they must concern
compatible entities or event stages, unless the question explicitly asks for
a comparison across different entities or events.  A valid set need not contain
one span per role or the same number of spans as another valid set.

The strict multi-video condition in
Equation~\ref{eq:strict-multivideo} is corpus-wide.  It is not established
merely because the annotated reference set contains several videos; every
single-video solution must remain insufficient.  \cmn{We verify this condition
for the 299 items whose annotated evidence spans at least two videos; the
remaining 101 items require multiple evidence spans within a single video.}  The corresponding
verification procedure is described in
Appendix~\ref{app:necessity-verification}.
When a seemingly sufficient prediction lacks a matching annotation, the
question is rewritten to rule it out.

\subsection{Solver-Side Contract and Evidence Semantics}

The solver receives none of $a^*$, $\mathcal{E}^{*}$,
$\mathcal{C}^{*}$, or $\mathcal{S}^{*}$.  The evaluated agent instantiates
$\operatorname{Decompose}(q)$ as an immutable, query-only working contract
created before retrieval.  It specifies material answer components, locally
observable evidence roles and their visible requirements, and any required
cross-role compatibility or joint relation.  This contract serves as a search and
verification scaffold.

The ledger $\mathcal{L}_t$, retained set $P_t$, and unresolved requirements
$U_t$ in Section~\ref{sec:problem} have distinct semantics.  A retrieved
candidate or predicted timestamp is only an address into the corpus.  It enters
$\mathcal{L}_t$ only after the corresponding visual content has been exposed
to the solver, and it enters $P_t$ only when that observation establishes a
concrete visible fact.  Consequently, retrieving a candidate does not itself
resolve a requirement, and a failed retrieval does not show that the required
evidence is absent from the corpus.  An unresolved requirement may be removed
only by an evidence-backed state update.

Local and set-level judgments are also separated.  A local observation may
establish the visible requirements of one evidence role.  Cross-video identity,
event compatibility, joint comparison or change, and complete answer support
must instead be assessed over the retained evidence set.  In the evaluated
agent, a local role is bound either to one observed member or to several
observations from the same source video.  A single role is never completed by
splitting its requirements across different videos; cross-video composition
occurs only among locally supported roles whose required compatibility and
joint relations are also verified.

\subsection{Video-Specific Support Constraints}

\paragraph{Static and temporal support.}
A displayed frame may establish an entity, attribute, visible state, count, or
spatial relation.  Motion, temporal order, transition, result, or continuity
requires temporal material in which the relevant stages are themselves
visible.  

\paragraph{Observation boundaries and continuity.}
An observation supports only what is visible within its displayed boundaries.
If a required initial state, action, identity link, or outcome lies outside the
window, the observation is partial.  Before--after observations may be
combined only when the relevant person, object, or event remains compatible;
a cut between two plausible states does not by itself establish continuity.

\paragraph{Viewpoint and cross-source composition.}
Screen-relative directions may reverse across front, rear, mirrored, or
opposing views, so directional conclusions require a shared coordinate system
or stable visual landmarks.  Likewise, observations from different products,
people, matches, performances, or news events cannot be combined merely
because they are topically or visually similar.  Cross-source composition is
valid only when the required compatibility is established or the question
explicitly requests comparison or aggregation across distinct instances.

\paragraph{Operational completion.}
The agent may stop when its retained evidence supports every material answer
component, all required compatibility and joint relations are satisfied, and
no material visual requirement remains unresolved.  This is an operational
stopping decision; benchmark evaluation independently scores the submitted
answer and cited spans.  If the budget is exhausted first or no legal productive action remains, the output is incomplete.
Since the current evaluation units are answerable,
such termination is treated as incomplete or premature.

\section{Benchmark Construction and Dataset Statistics}
\label{app:benchmark-details}

This appendix expands the construction of \textsc{DCI-QA}.  \textsc{DCI-Loc}
inherits its queries and reference moments from QVHighlights and uses the
candidate-corpus adaptation described in Section~\ref{sec:benchmark}.

\subsection{Corpus Composition and Freezing}

For each of 80 latent YouTube search seeds, we preserve an ordered top-20
search-result snapshot under a fixed collection configuration.  Before
inclusion, the returned videos are audited using pre-specified
technical-quality and near-duplicate criteria.  The original result identities
and ranks, collection time, video duration, availability state, and every
exclusion decision are retained so that filtering does not silently rewrite the
search context.  The seed and all construction metadata are hidden from the
evaluated solver.

The \textsc{DCI-QA} track contains 80 corpus groups, \cm{1,185} videos, and
400 questions.  Each question-level pool exposes \cm{14.8} usable video entries on
average.  The question-level pools are deliberately much longer than their
annotated evidence: \cm{domain-average pool duration ranges from 29 to 46
minutes, whereas the primary evidence occupies roughly 70 seconds per
question}. Details are shown in Table \ref{tab:dataset-statistics}.

\begin{table*}[t]
\centering
\caption{Domain composition and temporal scale of \textsc{DCI-QA}.
``Videos/Q'' is the mean number of videos in a question-level pool;
``Span dur.'' is the mean duration of one reference span;
``Evidence/Q'' is the mean total duration of the primary evidence set for one
question.  Pool duration is the total duration of the videos available for one
question, \cm{measured over the retained intervals exposed to the solver}, and
the range gives the shortest and longest question-level pools
within each domain.}
\label{tab:dataset-statistics}
\small
\setlength{\tabcolsep}{5pt}
\begin{tabular}{@{}lrrrrrrl@{}}
\toprule
Domain & Question share (\%) & Videos/Q & Spans/Q & Span dur. (s) &
Evidence/Q (s) & Pool mean & Pool range \\
\midrule
Fashion & \cm{25} & \cm{10.70} & \cm{3.01} & \cm{19.88} & \cm{59.82} & \cm{00:29:20} &
\cm{00:10:06--00:47:41} \\
Movie  & \cm{25} & \cm{15.80} & \cm{3.00} & \cm{21.32} & \cm{63.97} & \cm{00:40:35} &
\cm{00:17:38--00:58:30} \\
News    & \cm{25} & \cm{15.15} & \cm{3.00} & \cm{27.50} & \cm{82.51} & \cm{00:39:39} &
\cm{00:18:49--00:56:49} \\
Sports  & \cm{25} & \cm{17.60} & \cm{3.00} & \cm{23.87} & \cm{71.60} & \cm{00:46:12} &
\cm{00:15:41--00:58:45} \\
\bottomrule
\end{tabular}

\end{table*}

The questions are additionally tagged with twelve reasoning families for
stratified analysis, including cross-view alignment, fragmented chronology,
local--global spatial alignment, cross-context matching, object-state or
persistence reasoning, preparation-to-outcome verification, contradiction
resolution, and event identity or continuity.  These tags are never provided
to the solver.

\subsection{Question and Evidence Annotation}

After reviewing a corpus, annotators author five natural, open-ended questions
about visible actions, states, transformations, viewpoints, or relations
across its videos.  A question may name semantic event stages, but it does not
reveal the number of required observations, reference video identities,
timestamps, evidence-role names, or annotation order.  \cm{We remove questions
that require the audio track and questions that frame-level OCR alone can
answer.}

Following Appendix~\ref{app:extended-problem}, each accepted item stores a
reference answer, semantic evidence roles, localized reference spans, and one
verified primary evidence set.  A span records its source video, temporal
interval, and a concise statement of the directly visible fact.  Boundaries
retain enough context to verify the action, state, transition, or viewpoint
relation while excluding unrelated footage where possible.  \cm{The benchmark
contains 1,201 spans, or three per question, with a mean duration of 23.1
seconds and a range of 0.8 to 129.0 seconds.  Primary sets draw evidence from
one video for 101 questions, two videos for 106, and three videos for 193.}

Roles and spans need not be one-to-one: one span may establish several local
requirements, and a cross-video relation may require several spans.  The
solver is not given either count and may return a variable number of cited
spans within the evaluation limits.  Alternative support is handled by the
necessity and sufficiency checks below.

\subsection{Necessity and Shortcut Verification}
\label{app:necessity-verification}

A second annotator reviews each item with access to the question and complete
visual corpus but without the reference timestamps.  The review has three
parts.  First, the annotator verifies that the primary spans jointly establish
the reference answer and that removing a material role leaves at least one
answer component unsupported.  Second, the annotator searches the full pool
for a single video or a smaller evidence set that supports the complete
answer.  A distinct but
still strictly cross-video solution triggers a rewrite of the question.  Third,
the annotator checks that the question is well posed and that the required
entities, event stages, and relations can be distinguished visually.

We additionally audit six restricted-input conditions: question only; titles
and descriptions; speech transcripts; OCR text; thumbnails; and the strongest
single-frame candidate available from the visual index.  A correct model
prediction is treated only as a shortcut warning.  An item is revised or
removed only when a human reviewer can identify sufficient answer-bearing
information in the restricted input; chance guesses do not invalidate an
item.  \cm{The benchmark is available on Hugging Face at \dataurl.}

\section{Experiment Details}
\label{app:experiment}

This appendix records the implementation used
for the reported experiments.  

\subsection{Backbones and Model Roles}

We evaluate GPT-4o, GPT-5.2, GPT-5.4, GPT-5.5, GPT-5.6 Sol,
Gemini 2.5 Pro, Gemini 3 Pro, Gemini 3.5 Flash, and the
instruction-tuned Gemma 4 12B and 31B checkpoints.  Within a model arm, the
named backbone is used for all model-driven agent stages available in that
track: query-contract construction, search planning, Framesheet observation,
evidence-state updates, and citation localization; in \textsc{DCI-QA}, it also
performs evidence-set verification and answer synthesis.  The comparison
therefore changes the generative/VLM backbone while holding the tool
semantics, visual index, observation representation, and task-level search and
observation limits fixed.  Serving controls for hosted and local execution are
listed in Table~\ref{tab:model-implementation}.

\begin{table*}[t]
\centering
\caption{Model-facing implementation.  Paper model names identify the frozen
backbones used in the experiments.  The hosted and local arms share the same
agent logic and differ in serving and decoding controls.}
\label{tab:model-implementation}
\small
\setlength{\tabcolsep}{5pt}
\begin{tabular}{@{}p{0.15\textwidth}p{0.43\textwidth}p{0.35\textwidth}@{}}
\toprule
Execution & Backbones & Model-facing configuration \\
\midrule
Hosted
& GPT-4o; GPT-5.2; GPT-5.4; GPT-5.5; GPT-5.6 Sol; Gemini 2.5
  Pro; Gemini 3 Pro; Gemini 3.5 Flash
& Common multimodal Chat-Completions-compatible adapter; non-streaming,
  schema-constrained JSON; provider-default sampling and reasoning controls;
  512-token output cap. \\
\midrule
Local
& Gemma 4 12B IT; Gemma 4 31B IT
& Local vLLM serving in BF16 without quantization; temperature 0; thinking
  disabled; 8,192-token output cap. \\
\bottomrule
\end{tabular}

\end{table*}

All hosted backbones are accessed through the same project-level multimodal
adapter.  
Images are supplied as
visual content parts, and every model-facing stage requests a structured JSON
object that is validated locally.  Hosted requests do not explicitly set
temperature, top-$p$, a generation seed, or a reasoning budget and therefore
use the corresponding service defaults.

\subsection{Shared Cross-Modal Retrieval}

All model arms use a separate Gemini Embedding 2 backbone for candidate
retrieval.  It encodes model-generated textual search variants and
query-independent frame anchors in the same 3,072-dimensional multimodal
space; candidates are ranked by cosine similarity.  The evaluated backbone
controls query decomposition, search text, planning, observation, and evidence
use, so different models can still produce different retrieval trajectories
although the embedding encoder and frame index are shared.

The retrieval index contains no annotation text or gold-span proposals.  A
retrieved timestamp is only a candidate address: it is not counted as visually
observed until the corresponding timestamped frames are delivered to the VLM.
For \textsc{DCI-Loc}, the frozen index uses center-frame anchors at an
approximately two-second stride.  For \textsc{DCI-QA}, retrieval uses a
query-blind 1-FPS frame index, whereas actual local observation uses a
separate query-blind 3-FPS frame cache.  Thus, an embedding hit does not imply
that the model has inspected the denser observation frames.  The local Gemma
arms run their generative and visual-reasoning stages locally and use this same
text-query and cached frame-vector space for retrieval.

\subsection{Track-Specific Agent Configuration}

Both tracks use a visual-only interface: web search, titles, descriptions,
ASR transcripts, OCR indices, and generated captions are unavailable.  A
query-only contract is created once before retrieval and remains fixed during
the trajectory.  Round~0 uses controller-defined embedding/frame-skim retrieval; later
actions are selected only after the current observation has updated the public
evidence state.  Candidate spans remain unverified until their local windows
are materialized and inspected.

\begin{table*}[t]
\centering
\caption{Frozen interaction limits.  These values define what visual material
and search capacity are available to the agent.}
\label{tab:track-implementation}
\small
\setlength{\tabcolsep}{5pt}
\begin{tabular}{@{}p{0.12\textwidth}p{0.25\textwidth}p{0.11\textwidth}p{0.44\textwidth}@{}}
\toprule
Track & Visual observation & Round cap & Reproduction-critical limits \\
\midrule
\textsc{DCI-Loc}
& Timestamped Framesheets from one candidate video; single-video evidence
  state.
& 16
& Semantic observations contain at most 60 tiles.  After semantic
  finalization, citation localization may inspect at most five local windows,
  each no longer than 45 seconds, sampled at 2 FPS and capped at 90 tiles. \\
\midrule
\textsc{DCI-QA}
& Chronological Framesheets from the query-blind 3-FPS cache; cross-video
  retained-evidence set.
& 32
& Initial local windows are 6 seconds and expansions target 20 seconds.
  Global skim uses exactly 100 cached frames; a regional index uses at most
  100 frames; candidate reranking accepts 1--10 existing candidates.  Each
  retrieval action returns at most four candidates, the retained workspace
  holds at most eight evidence members, and set verification receives 2--4
  members. \\
\bottomrule
\end{tabular}

\end{table*}

In \textsc{DCI-Loc}, the agent searches a fixed ten-video cohort and maintains
a single-video evidence state.  Its final citation is produced only from
observed material, with a separate local refinement step used to tighten the
reported temporal boundary.  In \textsc{DCI-QA}, each retrieval or reranking
operation returns a bounded candidate batch; the selected candidate is
observed while unused candidates remain in a bounded workspace.  The local
observer updates only evidence roles within the current video.  Cross-video
compatibility, joint sufficiency, and final answer support are assessed over
the retained set.  Under full
corpus access, reference video identities and timestamps never enter the
controller.

\subsection{Local Serving and Execution Controls}

The Gemma checkpoints are served with vLLM in BF16, without weight
quantization, and with a 32,768-token model context.  Local campaigns run on
two NVIDIA H100 80GB GPUs, with one independent full-model replica per GPU and
deterministic case-level sharding; the replicas are not tensor-parallel.  All
generative and visual-reasoning stages run on the local replica, while the
shared Gemini Embedding 2 path remains responsible for text and frame
retrieval.

All model outputs are checked against strict stage-specific schemas.  A
malformed structured response may receive a bounded repair request, whereas
state reduction, candidate deduplication, overlap guards, budget accounting,
and final citation mapping are deterministic controller operations.  One
macro-round can consequently contain several model or retrieval calls.  The
semantic prompt bundle, preprocessing, index, and task-level controller policy
are frozen within each track.  Task-specific model coverage, repetition
counts, incomplete-run treatment, and analysis denominators are documented
with the result tables and in Appendix~\ref{app:evaluation-details}.

\section{Detailed Evaluation Protocol}
\label{app:evaluation-details}

This appendix expands the protocol summarized in
Section~\ref{sec:benchmark} and defines how the externally visible
trajectory is converted into the experiment results and human--agent comparison.  We
score only recorded candidates, displayed visual media, retained evidence, and
submitted citations.

\subsection{Observable Artifacts}

For evaluation item $i$, let $\mathcal{C}_i$ contain every candidate span
returned by a retrieval tool, including alternates that were never selected for
observation.  Let $\mathcal{D}_i$ denote the visual support actually displayed
to the solver: exact tile timestamps for a framesheet, or the actual playback
interval for a clip.  Let $R_i$ contain the active evidence retained in the
terminal public state, and let $P_i$ be the possibly empty set of temporal spans
submitted as final citations.  

These stages are ordered operationally but their reported metrics need not form
a strictly nested funnel.  For example, retained-video recall is a source-level
quantity, whereas displayed-span recall is a span-level quantity; a solver can
retain evidence from the correct video without having displayed a timestamp
inside the annotated moment.

The principal \textsc{DCI-Loc} analysis contains 500 QVHighlights queries.  A
planned run is indexed by query and repetition; the principal three-repetition
setting therefore has $N=1{,}500$ intention-to-treat (ITT) runs per model.  A
row evaluated with a different number of frozen repetitions uses its own
explicitly reported denominator and is not pooled with the three-repetition
rows.

To estimate confidence intervals for the \qa analysis, we additionally conduct three repeated evaluations on a fixed 115-case diagnostic subset. Results are shown in Appendix \ref{app:additional-results}.
For all ITT quality metrics, every planned unit remains in the denominator.
Missing artifacts, execution failures, unresolved runs, empty retained sets,
and empty citation sets contribute zero to rates and coverage measures.  For
the 1--5 answer score, a missing answer or incomplete valid judge pair receives
the scale floor of 1.  

\subsection{Temporal Overlap Quantities}

For a predicted interval $p=[s_p,e_p]$ and reference interval
$g=[s_g,e_g]$ from the same video, define
\begin{equation}
I(p,g)=\max\!\left(0,\min(e_p,e_g)-\max(s_p,s_g)\right).
\label{eq:eval-intersection}
\end{equation}
If $p$ and $g$ come from different videos, $I(p,g)=0$.  With
$\ell(p)=e_p-s_p$ and $\ell(g)=e_g-s_g$, we use
\begin{equation}
\operatorname{Cov}(p,g)=\frac{I(p,g)}{\ell(g)},\qquad
\operatorname{Prec}(p,g)=\frac{I(p,g)}{\ell(p)},
\label{eq:eval-coverage-precision}
\end{equation}
\begin{equation}
\operatorname{tIoU}(p,g)=
\frac{I(p,g)}{\ell(p)+\ell(g)-I(p,g)}.
\label{eq:eval-tiou}
\end{equation}
Coverage measures how much of the reference is recovered; precision measures
how much of the submitted interval is relevant; temporal IoU penalizes both
missing reference content and excessive temporal context.  All submitted and
reference intervals are required to have positive duration and to lie within
the source-video duration.

\subsection{\textsc{DCI-Loc} Source and Moment Measures}

For run $i$, let $v_i^*$ be the unique target video and let $G_i$ be its one or
more annotated moments.  The source-stage indicators are
\begin{align}
\operatorname{CandVHit}_i
&=\mathbf{1}\!\left[\exists c\in\mathcal{C}_i:
\operatorname{video}(c)=v_i^*\right],\\
\operatorname{FrameHit}_i
&=\mathbf{1}\!\left[\exists d\in\mathcal{D}_i,\,g\in G_i:
d\text{ is inside }g\right],\\
\operatorname{RetVHit}_i
&=\mathbf{1}\!\left[\exists r\in R_i:
\operatorname{video}(r)=v_i^*\right],\\
\operatorname{CiteVAcc}_i
&=\mathbf{1}\!\left[P_i\neq\emptyset\ \land\
\forall p\in P_i,\operatorname{video}(p)=v_i^*\right].
\label{eq:dci-loc-source-metrics}
\end{align}
Thus, candidate-video hit includes unobserved alternates, while FrameHit uses
only frame timestamps actually rendered to the model.  CiteV-Acc is strict: a
citation set containing both the target video and an additional wrong video
receives zero.  
RetV-Hit is retained as a trajectory diagnostic.

For final temporal localization, we consider every same-video pair
$(p,g)\in P_i\times G_i$ and choose the pair with maximum tIoU.  Ties are broken
in order by larger reference coverage, larger predicted-span precision, and a
fixed canonical span index.  If no same-video pair exists, all pair-level
quantities are zero.  Writing the selected pair as
$(p_i^\dagger,g_i^\dagger)$, we report
\begin{align}
\operatorname{R\mbox{-}Cov}
&=\frac{1}{N}\sum_i
\operatorname{Cov}(p_i^\dagger,g_i^\dagger),\\
\operatorname{mTIoU}
&=\frac{1}{N}\sum_i
\operatorname{tIoU}(p_i^\dagger,g_i^\dagger),\\
\operatorname{TIoU@.5}
&=\frac{1}{N}\sum_i
\mathbf{1}\!\left[
\operatorname{tIoU}(p_i^\dagger,g_i^\dagger)\geq0.5
\right].
\label{eq:dci-loc-temporal-metrics}
\end{align}
The main table reports mTIoU and TIoU@.5; the policy table additionally reports
R-Cov.  When citation refinement proposes a boundary outside the inspected
parent evidence episode, the deterministic controller falls back to the
pre-refinement citation.  This anchor-safe rule prevents an unsupported
refinement from receiving localization credit.

\subsection{\textsc{DCI-QA} Evidence-Set Measures}

Let $G_i$ be the reference spans for question $i$ and
$V_i^*=\{\operatorname{video}(g):g\in G_i\}$ the corresponding gold-video set.
For any artifact collection $A$, let $V(A)$ denote its unique source videos and
define gold-video recall
\begin{equation}
\operatorname{VR}_i(A)=
\frac{|V(A)\cap V_i^*|}{|V_i^*|}.
\label{eq:qa-video-recall}
\end{equation}
Candidate-video recall, retained-video recall, and cited-video recall are
$\operatorname{VR}_i(\mathcal{C}_i)$,
$\operatorname{VR}_i(R_i)$, and $\operatorname{VR}_i(P_i)$, respectively.
These are recalls rather than citation precision: an additional wrong video
does not erase a correctly recalled gold video.  They are therefore interpreted
alongside the bounded final evidence-set size and the temporal measures below.

Displayed-span hit recall is
\begin{equation}
\begin{gathered}
\operatorname{SpanHitRec}_i=\\
\frac{1}{|G_i|}\sum_{g\in G_i}
\mathbf{1}\!\left[\exists d\in\mathcal{D}_i:
d\text{ is displayed inside }g\right].
\end{gathered}
\label{eq:qa-span-hit}
\end{equation}
For Framesheets, this requires at least one exact displayed timestamp inside the
reference span; the surrounding candidate window and unsampled intervals do
not receive exposure credit.

Final citation spans are aligned to $G_i$ using a deterministic one-to-one
same-video matching.  Among all legal matchings, we first maximize total tIoU;
ties are broken by total reference coverage, total pair precision, matching
cardinality, and canonical indices.  One-to-one matching prevents a single
prediction from satisfying several reference spans.  Let $M_i$ be the selected
matching.  We report
\begin{align}
\operatorname{SpanCov}_i
&=\frac{1}{|G_i|}
\sum_{(p,g)\in M_i}\operatorname{Cov}(p,g),\\
\operatorname{Span@.5}_i
&=\frac{1}{|G_i|}
\sum_{(p,g)\in M_i}
\mathbf{1}\!\left[\operatorname{tIoU}(p,g)\geq0.5\right].
\label{eq:qa-span-metrics}
\end{align}
Unmatched reference spans contribute zero.  All DCI-QA evidence metrics are
computed per question and then macro-averaged over the fixed case set.

\subsection{Open-Ended Answer Quality}

Two fixed judges, GPT-5.4 and GPT-5.6 Sol, independently score each nonempty
candidate answer on an anchored 1--5 rubric for correctness and completeness.
The judges receive the question, reference answer, and candidate answer, but
not the evaluated system identity, retrieval scores, search trajectory, or gold
timestamps.  A score of 1 denotes an incorrect or nonresponsive answer and 5 a
complete and correct answer; intermediate scores represent increasing levels
of partial correctness or omission.

When both judgments are valid, the per-item score is
\begin{equation}
A_i=\frac{A_i^{(5.4)}+A_i^{(5.6)}}{2}.
\end{equation}
The combined-table ITT score is
\begin{equation}
    \begin{gathered}
        \operatorname{Ans\mbox{.-}5}_{\mathrm{ITT}}
=\frac{1}{N}\sum_{i=1}^{N}\widetilde{A}_i,
\\
\widetilde{A}_i=
\begin{cases}
A_i, & \text{complete valid judge pair},\\
1, & \text{otherwise}.
\end{cases}
    \end{gathered}
\label{eq:ans5-itt}
\end{equation}

\subsection{Search-Policy and Oracle Diagnostics}
\label{app:ablation-setting}

The ablation studies in Table \ref{tab:dci-loc-policy-ablation} and \ref{tab:dci-qa-evidence-access} modify the search or evidence-access condition while holding
the backbone, item set, observation format, downstream evidence pipeline, and
evaluation protocol fixed within each comparison.  The \textsc{DCI-Loc} policy
table uses Gemini 3 Pro on one frozen 500-case repetition.  The
\textsc{DCI-QA} oracle table uses GPT-5.5 on the fixed 115 cases, with each
intervention paired with the same Planner policy and repetition as its
Full-Corpus condition.  

\paragraph{Search-policy interventions.}
\textsc{Agent} is the full iterative reference policy: it searches the complete
ten-video cohort, observes retrieved regions, updates its evidence state, and
may redirect subsequent searches in response to the evidence already found.
\textsc{Uniform-Skim} replaces query-conditioned retrieval and search planning
with a fixed corpus-wide schedule.  It displays nine uniformly spaced frames
from each of the ten videos, for 90 frames in total, regardless of the query or
intermediate evidence.  These observations are then processed by the same
Observer, evidence-state update, retention, final-citation, and boundary-
refinement stages as the Agent.  

\textsc{One-Shot} is paired with the corresponding Agent run and reuses its
Round-0 candidate batch.  It observes only the top-ranked candidate and then
disables further candidate retrieval, inspection of lower-ranked candidates,
and evidence-conditioned replanning.  The ordinary evidence reduction, retained
evidence, final citation, and boundary-refinement stages remain active.
Consequently, One-Shot denotes a single candidate-search stage.  The Agent--One-Shot comparison therefore measures the value
of continuing the evidence-search loop after the same initial candidate,
whereas Agent--Uniform-Skim compares adaptive, query-conditioned search with a
fixed global-coverage policy.

\paragraph{Oracle interventions.}
\textsc{Oracle-Video} removes source discovery at the Controller level by
restricting the runtime index to the videos belonging to one acceptable
reference set.  It does not reveal the relevant timestamps or attach any
``gold'' label to the remaining videos.  The Agent must still search within
those videos, interpret the observed content, retain the relevant evidence, and
produce its final localization or answer.  In \textsc{DCI-Loc}, this normally
leaves one authorized target video; in \textsc{DCI-QA}, the restricted set may
contain multiple videos whose evidence must still be combined.

\textsc{Oracle-Span} additionally removes initial temporal-candidate discovery
by injecting the corresponding reference windows as ordinary, unverified
candidates.  For \textsc{DCI-Loc}, the Controller deterministically selects one
eligible reference span when several alternatives exist.  For
\textsc{DCI-QA}, it supplies all active localized spans in the selected
reference set, because the answer may require complementary evidence from
several videos.  The model is not given the reference answer, annotation
provenance, or an indication that these candidates are correct.

All downstream visual observation, evidence-state update, retention, set
verification where applicable, citation construction, boundary refinement, and
answer generation remain unchanged.  Oracle rows may therefore still fail
because of visual misunderstanding, evidence loss, incorrect composition or
refinement, malformed output, or execution failure.  Accordingly,
Oracle-Video minus the full-search condition diagnoses the cost of source
discovery, while Oracle-Span minus Oracle-Video diagnoses the additional cost
of temporal candidate discovery.  

\subsection{Round-Wise Evidence Use and Efficiency}

Let $\mathcal{D}_{i,\leq t}$ be all media displayed through macro-round $t$.
For \textsc{DCI-Loc}, cumulative any-gold exposure is
\begin{equation}
\operatorname{Seen}_i(t)=
\mathbf{1}\!\left[\exists d\in\mathcal{D}_{i,\leq t},\,g\in G_i:
d\text{ hits }g\right].
\label{eq:roundwise-seen}
\end{equation}
The cumulative cited curve records whether a run has terminated by round $t$
with a final anchor-safe citation that hits a gold moment.  It does not imply
that a final citation was emitted at every preceding round.  Once a run stops,
its terminal state is carried forward to the maximum plotted round; unresolved 
runs remain zero.  

For \textsc{DCI-QA}, the round-wise seen and cited curves are case-level
\emph{any-gold} indicators.  Seen becomes positive once an exact displayed
timestamp falls inside any reference span.  Cited becomes positive once the
final citation set contains a same-video interval with positive temporal
overlap with any reference span.  These curves are intentionally different
from the unit-level video and span recalls in the tables; a case can have
any-gold success while recovering only one of several required evidence
members.

Rounds is the mean number of semantic macro-rounds in the table-specific
trajectory population.  Time is the median active per-case
latency, excluding outer campaign queueing but including the model, retrieval,
and deterministic controller work performed while the case is active.  Hosted
and local times are operational end-to-end measurements without hardware
normalization.  
\subsection{Human Reference and Figure Construction}
\label{app:human-protocol}
The \textsc{DCI-Loc} panels in
Figure~\ref{fig:gap-decomposition} use measured per-case trajectories on the
50 items shared by the human pilot and agent runs.  Agent repetitions are first averaged within case, after
which the Agent Average is a model-macro average over the hosted systems.  The
Best Agent is GPT-5.5, the system with the highest Ans.-5 in
Table~\ref{tab:dci-main-combined}.

The shared-subset localization panel uses three endpoint measures.  Correct
video tests whether the final citation selects the target source.  Final
PointLoc treats the final two-second point anchor as correct when it overlaps
any same-video reference moment.  Boundary IoU@.5 applies the full submitted
boundary and requires temporal IoU of at least 0.5.  The cumulative cited
curve and the cited segment of the exposure--citation partition use the
PointLoc criterion, whereas the seen curve uses only media actually displayed
to the solver.

The DCI-QA funnel panel uses measured matched-protocol human aggregate points.
For agents, the hosted systems are first averaged within case and then
aggregated over cases for the Agent Average; the Best Agent remains the fixed
GPT-5.5 arm.
The round-wise DCI-QA human seen/cited curve is based on a measured 50-case human
trace.  
The human and agent interfaces are functionally matched
in source selection, bounded local inspection, evidence submission, and
stopping.

\subsection{Uncertainty, Paired Comparisons, and Alternate Evidence}

We use 10,000 percentile-bootstrap resamples.  DCI-Loc full-set and policy
comparisons resample fixed ten-video cohorts so that queries sharing a
candidate pool are not treated as independent.  DCI-QA analyses resample the 115-subset.  
Pairwise intervention effects are computed within
item before cluster resampling and are reported in percentage points for rate
or coverage metrics.  \apre{Results with 95\% paired intervals in Table \ref{tab:dci-main-combined}, \ref{tab:dci-loc-policy-ablation} and \ref{tab:dci-qa-evidence-access} are shown in Appendix \ref{app:additional-results}.}

\section{Prompt and Tool Specifications}
\label{app:agent-tools}

This appendix describes the final prompt and tool contracts used by the
reported agents.  
The reported runs use
timestamped frame sheets.
The system is a controller-mediated evidence agent.
Model calls perform bounded semantic judgments, while
the controller owns identifiers, temporal mapping, persistent state, tool
execution, budgets, stopping gates, and final citation materialization.
Retrieval outputs are always candidate locations and do not become evidence
until their local visual content has been displayed and evaluated.

\subsection{Controller--Model Responsibility Boundary}

Every model-facing stage receives only the information needed for its assigned
role and returns a schema-constrained JSON object.  A common wrapper preserves
the Controller-specified order of text fields and images, requests a structured
response, parses it locally, and applies both schema and semantic validation.
JSON mode alone is not treated as sufficient validation: the Controller checks
all referenced identifiers, enumerated values, media coordinates, evidence
bindings, tool limits, and stage-specific invariants.

A rejected structured output may receive a bounded repair request containing
the rejected draft and concise validation issues.  Repair is used only to
correct an invalid response; each valid output is used as generated, without
resampling or best-of-$n$ selection.  If repair remains unsuccessful, the stage fails
closed or follows its deterministic fallback.  

\begin{table*}[t]
\centering
\caption{Division of responsibility among model prompts and deterministic
Controller operations.  Each model stage is restricted to one semantic task;
structural state transitions and final evidence identifiers remain
Controller-owned.}
\label{tab:prompt-stage-boundaries}
\small
\setlength{\tabcolsep}{4pt}
\begin{tabular}{@{}p{0.13\textwidth}p{0.25\textwidth}
p{0.27\textwidth}p{0.27\textwidth}@{}}
\toprule
Stage & Model input & Model output & Controller-enforced boundary \\
\midrule
Query contract
& Question only
& Stable visual requirements and their structural dependencies
& No media, retrieval result, reference annotation, timestamp, or guessed
answer is available.  The validated contract is generated once and then
remains immutable. \\
\midrule

Local observer
& Question, immutable contract, compact state for one current video, and one
ordered local frame observation
& Directly covered requirements, a calibrated support label, a concise visible
fact, and any unresolved local gap
& Cannot select a tool, inspect another source, alter the global state, or
declare cross-video sufficiency. \\
\midrule

Search planner
& Compact retained evidence, active needs, candidate workspace, bounded
progress summaries, and remaining limits
& Exactly one tagged tool action, or a terminal proposal
& Does not receive visual media and cannot write evidence updates, citations,
or the final answer.  All IDs, ranges, and tool arguments are revalidated. \\
\midrule

Evidence-set verifier
(\textsc{DCI-QA})
& An ordered bundle of 2--4 inspected evidence members and a
Controller-generated verification blueprint
& Role, compatibility, joint-reasoning, and answer-component judgments
& Cannot add, remove, reorder, or replace evidence members, invoke a tool, or
write the user-facing answer. \\
\midrule

Answer synthesizer
(\textsc{DCI-QA})
& Materialized set verification, selected evidence records, and legal claim
slots
& Answer text, slot-bound claims, unresolved-component statements, and
limitations
& Claim IDs, direct versus joint support, evidence IDs, answer status, and
citations are owned and materialized by the Controller. \\
\midrule

Citation refiner
(\textsc{DCI-Loc})
& Finalized semantic state, preliminary evidence anchors, and an expanded
ordered frame observation
& Goal-local frame ranges and visible boundary descriptions
& Cannot change semantic support, the selected source video, the answer, or
the search state; invalid refinements fall back to preliminary citations. \\
\bottomrule
\end{tabular}

\end{table*}

All evaluated backbones use the same prompt text, output schemas, and
controller policy within a track.  The query contract is nevertheless
model-generated: each backbone produces one contract per item, which is then
frozen and reused wherever that model and item appear in repeated or diagnostic
conditions. 

\subsection{\textsc{DCI-Loc} Prompt Pipeline}

\paragraph{Query contract.}
Before retrieval, the contract prompt converts the query into a small set of
visual goals.  Each goal contains one to four directly observable requirements,
is marked as either a claim or an answer-bearing goal, and specifies whether
its requirements must be established in one observation or may be accumulated
across observations from the same video.  Optional lightweight dependencies
preserve an explicitly required order, event relation, or visible entity
binding.  Nonvisual conditions such as intention, external identity, or
unstated background facts remain auxiliary limitations; they neither become
visual search needs nor authorize visual guessing.

The reported localization agent uses single-video completion.  It may inspect
many sources, but finalization requires one video thread to satisfy every
material visual goal, answer requirement, and binding.  Evidence from different
videos is never combined to complete a localization item.

\paragraph{Local visual observation.}
The frame Observer receives the original query, the compact evidence thread for
the current video, a Controller-authored frame locator, and one ordered frame
batch.  It returns only the current observation's delta.  For each materially
addressed goal, it selects one of
\texttt{support}, \texttt{partial\_support}, \texttt{ambiguous}, or
\texttt{not\_verified}; lists the requirement IDs directly visible in the
frames; states one concise visible fact; and, when evidence is incomplete,
specifies an unresolved gap.

The Observer additionally localizes each covered requirement to visible
one-based frame indices.  It does not output absolute timestamps.  The
Controller deterministically maps accepted frame ranges to source-video
coordinates and rejects ranges that reference unseen frames or requirements
not covered by the semantic judgment.

The frame prompt explicitly distinguishes sampled states from continuous
video.  A static frame may establish an entity, state, count, or spatial
relation.  An ordered batch may establish a sampled state difference when the
subject remains identifiable, but sparse endpoints do not by themselves
establish unseen motion, uninterrupted continuity, mechanism, or causality.
Failure to verify the current candidate is local to that observation and is not
interpreted as evidence that the target is absent from the corpus.

\paragraph{State update and search planning.}
A deterministic reducer merges valid deltas into a bounded per-video evidence
thread.  Requirements marked as requiring one observation cannot be completed
by combining unrelated frame batches.  Requirements marked as same-video may
accumulate across observations from that video, but this aggregation does not
license an unseen transition or identity relation.

Uncovered requirements create persistent needs keyed by the video thread, goal,
and requirement IDs.  A need remains active until its requirements are
supported or explicitly superseded; it does not disappear merely because a
later Observer omits it.  The planner receives these active needs together with
bounded evidence-thread summaries, unobserved candidate alternates, recent
progress, and the remaining tool limits.  It returns exactly one legal action,
\texttt{finalize}, or \texttt{insufficient}.  Round~0 is fixed
embedding-based discovery and does not invoke the planner.
A planner proposal to finalize is accepted only when one video thread satisfies
the complete contract and has no unresolved material need.  

\paragraph{Citation-boundary refinement.}
After semantic finalization, a separate refiner examines an expanded frame
window around each preliminary evidence anchor.  It identifies the shortest
defensible frame range containing the visible event and marks whether the event
appears complete or truncated at either observation boundary.  The refiner is
not allowed to revise which video or event supports the query.  Its proposed
range must remain connected to the preliminary inspected episode; otherwise
the Controller retains the preliminary citation.  The observation and
refinement limits are those reported in
Table~\ref{tab:track-implementation}.
There is no separate free-form answer prompt for \textsc{DCI-Loc}.  Any
answer-bearing value is obtained during local observation, and the final
source, answer fields, and citations are projected deterministically from the
validated terminal state.

\subsection{\textsc{DCI-QA} Prompt Pipeline}

\paragraph{Query contract.}
The QA contract separates the question into one to four material answer
components and two to four locally observable evidence roles.  Each role
describes one coherent local instance that can normally be inspected in a
short temporal window and contains one to four visually discriminative
requirements.  The contract additionally records only those cross-role
compatibility relations and joint reasoning requirements that are material to
the requested answer.

Compatibility and joint reasoning are kept separate.  Compatibility determines
whether locally supported members may be composed---for example, whether they
show the required object, event stage, viewpoint relation, or intended
comparison group.  Joint reasoning specifies a conclusion that genuinely
requires several roles, such as a visible comparison or before--after relation.
The contract also maps every answer component to the roles and joint relations
required to support it.  Orphan roles and generic ``relevant evidence'' roles
are rejected by validation.

\paragraph{Initial retrieval and local observation.}
The default Full-Corpus trajectory constructs one role-focused retrieval lane
for each evidence role.  These lanes use the shared embedding index and are
assembled in role order with source diversity, yielding at most four initial
candidates.  

A selected candidate is materialized as a chronological local observation from
the query-blind frame cache.  One observation may be divided into up to three
frame sheets for input capacity, but all sheets belong to one source video and
one continuous temporal window and jointly produce at most one
\emph{EvidenceMember}.  The Observer reads them in Controller-specified order
and returns role-level support labels, covered requirement IDs, a concise local
fact, an unresolved gap, and a small number of directly visible linking cues.

Linking cues describe observable appearance, object, setting, event-stage, or
viewpoint features that may later help compare members.  They do not assert
that two videos depict the same person, object, garment, or event.  The local
Observer is prohibited from deciding cross-video compatibility, causal
relations, joint sufficiency, or the final answer.

\paragraph{Evidence-set verification.}
\cmn{Once the retained pool can form a candidate set from at least two
qualifying evidence spans}, the Controller constructs an ordered set of 2--4 EvidenceMembers.  It
also constructs a verification blueprint enumerating every structurally legal
grounding option for:

\begin{itemize}
    \item each local evidence role;
    \item each required compatibility relation;
    \item each joint reasoning requirement; and
    \item each material answer component.
\end{itemize}

The verifier inspects only this fixed visual bundle, the locally verified
records, the immutable contract, and the legal blueprint.  It cannot complete
one local role by splitting its requirements across videos.  It assigns each
target a calibrated semantic status and chooses only among the grounding
options provided for that target.  The set is \texttt{sufficient} only when all
required roles, compatibility relations, joint relations, and answer
components are supported.  Otherwise it returns a structured partial,
ambiguous, or incompatible diagnosis and identifies the remaining gaps.

The model does not directly select final evidence IDs.  The Controller
materializes the selected grounding options into ordered evidence records,
which prevents the verifier from inventing members, silently changing their
order, or citing media outside the evaluated bundle.

\paragraph{Answer generation and calibration.}
After search terminates, the Controller constructs a claim blueprint from the
materialized set verification.  Each legal claim slot is marked as either
direct, grounded in one EvidenceMember, or joint, grounded in an explicitly
supported multi-member relation.  The answer synthesizer writes only the
natural-language content assigned to these slots.

Without a supported joint slot, the model may juxtapose separate local
observations but may not claim cross-video identity, comparison, change,
mechanism, or causality.  It also cannot output evidence IDs, claim IDs, answer
status, or citation arrays.  The Controller determines whether the output is
complete, qualified partial, or insufficient; assigns the identifiers; and
maps the selected EvidenceMembers to temporal citations.  If material answer
components remain unresolved, the answer must state the limitation explicitly
rather than filling it from the question or outside knowledge.

\paragraph{Planning and bounded local recovery.}
The QA planner is called only after the latest observation and any eligible
Controller-owned local recovery have updated the state.  Before a new global
planning call, the Controller may perform at most two consecutive local
actions, such as consuming a targeted workspace alternate, expanding the
current temporal window, or searching a bounded fresh region of the same
video.  A no-progress local observation immediately forces planner re-entry;
even after progress, the local tranche ends after its bounded allowance.

The planner receives retained EvidenceMembers, the latest verifier diagnosis,
active needs, bounded workspace and progress summaries, remaining limits, and
an authoritative list of legal or exhausted actions.  It does not inspect
frames itself.  It must select one action with expected information gain and
may finalize only when the latest set verification is
\texttt{sufficient}.  Retrieved or observed video count alone is not a
finalization condition.

\begin{table*}[t]
\centering
\caption{Normalized planner-tool contracts, grouped by their effect on the
search space. The \emph{output} and \emph{search-space effect} columns
distinguish corpus discovery, within-source localization, candidate
prioritization, observation, and temporal expansion. Retrieval and selection
outputs remain candidate locations; only ordinary Observer processing and the
subsequent state update can promote an inspected window to evidence.
QA denotes \textsc{DCI-QA}, while Both denotes \textsc{DCI-Loc} and
\textsc{DCI-QA}. Track-specific durations, frame capacities, and macro-round
limits are reported in Table~\ref{tab:track-implementation}.}
\label{tab:detailed-tool-contracts}
\footnotesize
\setlength{\tabcolsep}{3pt}
\renewcommand{\arraystretch}{1.12}

\begin{tabular}{@{}
>{\raggedright\arraybackslash}p{0.120\textwidth}
>{\raggedright\arraybackslash}p{0.160\textwidth}
>{\raggedright\arraybackslash}p{0.240\textwidth}
>{\raggedright\arraybackslash}p{0.130\textwidth}
>{\raggedright\arraybackslash}p{0.200\textwidth}
>{\centering\arraybackslash}p{0.055\textwidth}
@{}}
\toprule
Tool
& Scope and input
& Primary operation
& Output
& Search-space effect
& Track \\
\midrule

\multicolumn{6}{@{}l}{\textit{Corpus-level discovery}}\\[-2pt]
\cmidrule(lr){1-2}
Atomic embedding
& Full corpus; requirement-focused visual query
& Embeds the query and cached frames in the shared space, retrieves frame
anchors, and temporally deduplicates the results.
& At most four candidate spans
& May introduce previously unseen source videos and temporal regions.
& Both \\

\addlinespace[3pt]
\cmidrule{1-6}
Global skim
& Full corpus; fixed sheet of exactly 100 query-blind cached frames
& Asks the VLM to select promising visual anchors from the
Controller-selected sheet.
& At most four candidate anchors
& Selects only from the fixed sheet; it performs no query-conditioned
resampling.
& QA \\

\addlinespace[5pt]
\midrule
\multicolumn{6}{@{}l}{\textit{Within-source localization}}\\[-2pt]
\cmidrule(lr){1-2}
Regional VLM index
& One known video; one approved range containing at most 100 cached frames
& Projects frames from the range and applies hierarchical visual selection
when needed.
& At most four local candidate anchors
& Refines only within the approved range; it cannot scan another video or an
exhausted overlapping range.
& Both \\

\addlinespace[3pt]
\cmidrule{1-6}
Candidate rerank
& Between one and ten existing localized candidates
& Generates a precise visual criterion and scores representative frames from
the supplied candidates.
& Reordered candidate list
& Changes candidate order only; it creates no new video, region, or candidate.
& Both \\
\midrule
\addlinespace[5pt]
\multicolumn{6}{@{}l}{\textit{Candidate inspection and contextualization}}\\[-2pt]
\cmidrule(lr){1-2}
Observe known candidate
& Existing workspace; one previously returned, unobserved candidate
& Selects the candidate and passes its local window to the ordinary Observer.
& One local window submitted for observation
& Consumes an existing candidate without performing new retrieval or expanding
coverage.
& Both \\

\addlinespace[3pt]
\cmidrule{1-6}
Expand current observation
& Current source and temporal observation
& Extends the current window to recover a nearby action endpoint, outcome, or
missing context.
& One expanded candidate window
& Adds adjacent time from the same source only; it does not directly update the
evidence state.
& Both \\

\bottomrule
\end{tabular}

\end{table*}

\subsection{Tool Contracts}

The low-level tools normalize common coarse-to-fine video-search primitives
into a controlled and auditable action space shared across backbones.  Every retrieval or reranking operation returns at most four ranked
\emph{CandidateSpan}s.  Each candidate records a source, temporal proposal,
rank or retrieval signal, and tool provenance, but remains unverified until its
local window is observed.

Algorithm~\ref{alg:evidence-search} presents the union of normalized tool
families.  The reported tracks expose different subsets, summarized in
Table~\ref{tab:detailed-tool-contracts}.  In particular, the adaptive
Global-Skim action is enabled in the QA agent.

\paragraph{Atomic embedding.}
The retrieval text is derived from an unresolved visual requirement while
preserving material subject, object, event-stage, and viewpoint constraints.
The evaluated backbone determines this search text, but all arms use the shared
embedding encoder and cached frame space described in
Appendix~\ref{app:experiment}.  A high cosine score is therefore navigation
evidence only; it provides no support label and no visual-exposure credit.

\paragraph{Global skim.}
Global Skim differs from the non-adaptive \textsc{Uniform-Skim} baseline.
It is one planner-selected visual discovery action targeted to a current
evidence need, whereas \textsc{Uniform-Skim} follows a fixed observation
schedule.  The skim model sees request-local frame indices and images, not gold
locations or annotations.  Even a visually strong selected frame is converted
into a local temporal candidate and passed to the Observer; sparse skim frames
cannot directly establish a full action, continuity, or cross-video relation.

\paragraph{Regional indexing and reranking.}
Regional indexing is permitted only when the Controller already knows a source
video and approves a fresh temporal range.  Candidate reranking has a narrower
scope: it receives only existing candidate IDs and cannot search outside them.
This distinction prevents a nominal reranker from becoming an unreported
whole-video scanner.

\paragraph{Workspace consumption and expansion.}
Unused results from a candidate batch enter a bounded workspace.  A relevant
alternate can later be consumed with
\texttt{observe\_known\_candidate}, avoiding a repeated retrieval call.
Expansion is permitted only around the exact current media item and is rejected
when it would merely repeat a high-overlap, no-progress observation.

\subsection{Persistent State and Execution Guards}

The Controller maintains three bounded structures in addition to retained
evidence.

\paragraph{Need registry.}
An unresolved local requirement creates a persistent need keyed primarily by
its video context, role or goal ID, and requirement IDs.  Rephrasing the same
gap does not create a new need.  Covering the relevant requirements resolves
the need; omission from a later model delta does not.

\paragraph{Candidate workspace.}
The workspace stores ranked observed and unobserved CandidateSpans, their source
actions, and temporal provenance.  Candidates are deduplicated using source
video and temporal overlap rather than transient media identifiers.  The
highest-priority valid candidate is materialized, while alternates remain
available for later observation.

\paragraph{Progress memory.}
The Controller distinguishes retrieval outcomes from evidence outcomes.
A tool may return \texttt{candidate\_found}, but the subsequent observation can
still be \texttt{not\_verified}, \texttt{ambiguous}, duplicate, or otherwise
produce no evidence gain.  Positive progress requires a new covered
requirement, a narrower unresolved gap, a newly supported role, or a stronger
set-level state.  Only bounded summaries of recent attempts are exposed to the
planner; raw unbounded histories are not appended to every prompt.

Overlap-aware guards reject or force replanning after a repeated action targets
the same need, source, and substantially overlapping temporal region without
new coverage.  The planner must then change the source, temporal cluster,
visual criterion, or tool family, or terminate when no legal productive action
remains.  A new media identifier cannot bypass this check.

The deterministic stopping gates remain track-specific.  Localization requires
one complete video thread.  Cross-video QA requires a verifier-approved
sufficient evidence set with the required source diversity and all material
answer components supported.  Oracle conditions change only which candidate
partition or temporal windows the Controller may expose; they do not weaken
Observer, verifier, citation, or answer-generation requirements.

\subsection{Prompt Freezing and Reproducibility}

The reported comparisons use one frozen prompt, schema, and Controller bundle
per track.  For reproducibility, we record the verbatim prompt text, output
schema, cryptographic hash, tool allowlist, media ordering and index
conventions, explicit versus default decoding controls, bounded repair policy,
and Controller configuration associated with each reported run.  

\section{Additional Results}
\label{app:additional-results}

This appendix complements the point estimates in
Tables~\ref{tab:dci-main-combined}--\ref{tab:dci-qa-evidence-access} with
cluster-bootstrap 95\% confidence intervals.

\subsection{Main result}

Table~\ref{tab:dci-main-combined-three-repetition-ci} summarizes uncertainty
across evaluated models for both \loc and \qa, extending Table \ref{tab:dci-main-combined}.

\begin{table*}[t]
\centering
\caption{Cluster-bootstrap 95\% confidence intervals for the main
\textsc{DCI-Loc} and \textsc{DCI-QA} results.  Each entry reports only the
lower and upper endpoints; the corresponding point estimates appear in
Table~\ref{tab:dci-main-combined}.  All endpoints are reported to three
significant figures.  Intervals use 10,000 deterministic resamples over 50
\textsc{DCI-Loc} cohorts or 52 \textsc{DCI-QA} construction seeds, with
repetitions kept inside their query cluster.  Dashes denote unavailable
results.  Bold and underlined intervals correspond to the best and second-best
point estimates, respectively, among evaluated agents in
Table~\ref{tab:dci-main-combined}.}
\label{tab:dci-main-combined-three-repetition-ci}
{\small
\setlength{\tabcolsep}{2.5pt}
\begin{tabular*}{\textwidth}{@{\extracolsep{\fill}}lccccccc@{}}
\toprule
\multicolumn{8}{@{}l}{\textbf{\textsc{DCI-Loc}: Temporal Localization}} \\
\midrule
Model & CandV-Hit & FrameHit & CiteV-Acc & mTIoU & TIoU@.5 &
Rounds $\downarrow$ & \shortstack{Time (min)\\$\downarrow$} \\
\midrule
GPT-4o            & 80.8--86.1 & 66.7--73.5 & 17.7--24.5 & 7.50--10.8  & 6.70--10.4  & \textbf{3.82--4.01} & \textbf{1.29--1.34} \\
GPT-5.2           & 86.0--90.7 & 76.5--82.1 & 33.8--42.6 & 17.2--22.5 & 17.6--23.8 & 5.93--6.44 & 2.25--2.44 \\
GPT-5.4           & 85.3--90.1 & 74.0--80.5 & 45.0--54.3 & 21.3--26.6 & 21.9--27.7 & \underline{4.36--4.60} & 1.55--1.63 \\
GPT-5.5       & 86.9--91.1 & 78.9--83.9 & \underline{55.5--63.0} & \underline{23.8--27.9} & \underline{23.9--28.9} & 6.03--6.59 & 3.11--3.51 \\
GPT-5.6 Sol       & \textbf{88.9--93.1} & \textbf{82.0--87.1} & 52.9--61.3 & 23.1--28.3 & 23.0--29.3 & 5.42--5.84 & 1.91--2.09 \\
Gemini-2.5 Pro    & 81.9--87.3 & 68.9--75.7 & 54.8--62.7 & 21.9--26.4 & 21.7--27.3 & 4.84--5.15 & 3.88--4.25 \\
Gemini-3 Pro      & 82.1--87.3 & 70.5--77.3 & 54.8--61.6 & 21.5--26.1 & 21.8--27.3 & 4.94--5.30 & 3.95--4.39 \\
Gemini-3.5 Flash  & \underline{89.1--93.4} & \underline{81.2--85.9} & \textbf{67.5--74.6} & \textbf{26.9--31.8} & \textbf{25.3--31.0} & 5.10--5.61 & 2.38--2.68 \\
Gemma 4 12B       & 71.8--78.6 & 62.6--70.5 & 30.9--36.9 & 10.2--13.6 & 9.60--14.1  & 5.77--6.06 & \underline{1.51--1.59} \\
Gemma 4 31B       & 68.9--76.4 & 62.2--70.1 & 46.8--55.0 & 17.3--22.2 & 16.9--23.1 & 5.01--5.31 & 2.32--2.58 \\
\bottomrule
\end{tabular*}

\par\medskip

\begin{tabular*}{\textwidth}{@{\extracolsep{\fill}}lcccccccc@{}}
\toprule
\multicolumn{9}{@{}l}{\textbf{\textsc{DCI-QA}: Cross-Video Evidence QA}} \\
\midrule
Model & \shortstack{CandV\\Rec} & \shortstack{SpanHit\\Rec} &
\shortstack{RetV\\Rec} & \shortstack{CiteV\\Rec} & SpanCov & Ans.-5 &
Rounds $\downarrow$ & \shortstack{Time (min)\\$\downarrow$} \\
\midrule
GPT-4o            & 39.3--54.7 & 4.40--10.9  & 8.70--19.0  & 7.30--15.1  & 1.30--4.00  & 1.13--1.22 & \textbf{1.89--3.02}  & \textbf{2.48--2.69}  \\
GPT-5.2           & 56.4--70.4 & 30.1--43.9 & 47.7--61.5 & 32.7--46.3 & 7.10--11.2 & 1.77--2.03 & 9.04--11.4 & 5.70--7.46  \\
GPT-5.4           & 60.7--73.7 & \underline{29.5--41.7} & 49.5--63.5 & 34.6--48.1 & 5.80--9.60  & 2.20--2.53 & 8.32--10.0 & \underline{3.31--4.14}  \\
GPT-5.5       & \textbf{63.9--78.1} & 32.3--44.9 & \underline{54.3--68.3} & 37.8--51.6 & \underline{7.40--12.6} & \textbf{2.42--2.75} & 7.84--10.3 & 4.27--7.42  \\
GPT-5.6 Sol       & \underline{63.2--76.6} & \textbf{35.3--47.6} & \textbf{56.5--69.6} & \textbf{39.3--54.4} & \textbf{8.20--13.0} & \underline{2.30--2.65} & 9.03--11.1 & 4.04--5.81  \\
Gemini-2.5 Pro    & 57.7--72.1 & 25.5--36.6 & 45.8--58.0 & 32.9--45.4 & 5.80--10.9 & 1.75--2.08 & 7.40--8.95  & 8.75--11.1 \\
Gemini-3 Pro      & 61.5--74.4 & 25.7--37.5 & 45.3--58.2 & 30.9--42.7 & 5.90--10.4 & 1.73--2.02 & 7.84--9.84  & 7.59--10.4 \\
Gemini-3.5 Flash  & 57.5--71.2 & 28.5--39.7 & 38.8--52.6 & \underline{40.6--55.2} & 7.30--12.0 & 2.02--2.33 & 9.58--11.6 & 5.81--7.62  \\
Gemma 4 12B       & 44.2--58.9 & 12.8--23.8 & 24.4--37.6 & 21.7--34.6 & 3.40--7.80  & 1.35--1.55         & \underline{2.91--3.64}  & 4.51--4.96  \\
Gemma 4 31B       & 55.4--69.9 & 22.5--34.4 & 41.0--55.3 & 37.8--50.8 & 6.50--10.1 & 1.67--1.98         & 5.61--6.71  & 7.30--8.75  \\
\bottomrule
\end{tabular*}}

\end{table*}

\subsection{Ablation studies}

Table~\ref{tab:dci-loc-policy-ablation-ci} and \ref{tab:dci-qa-evidence-access-ci}  compare fixed-schedule,
single-step, and adaptive search with privileged video- and span-access
diagnostics, extending Table \ref{tab:dci-loc-policy-ablation}, \ref{tab:dci-qa-evidence-access}.  The controlled settings are detailed in
Appendix~\ref{app:ablation-setting}.

\begin{table*}[t]
\centering
\caption{\textsc{DCI-Loc} search-policy and evidence-access diagnostics
(Gemini 3 Pro), using queries across three repetitions (1,500 ITT
slots/row).  Entries report only the endpoints of 50-cohort bootstrap 95\%
confidence intervals, each to three significant figures. Bold and underlined intervals follow the best and second-best
underlying point estimates, respectively, across rows.}
\label{tab:dci-loc-policy-ablation-ci}
{\small
\setlength{\tabcolsep}{4pt}
\begin{tabular*}{\textwidth}{@{\extracolsep{\fill}}lccccc@{}}
\toprule
Policy & FrameHit $\uparrow$ & FinalV $\uparrow$ & R-Cov. $\uparrow$ &
IoU@.5 $\uparrow$ & Rounds $\downarrow$ \\
\midrule
Uniform-Skim & \underline{91.7--96.1} & 42.2--50.2 &
\underline{33.5--40.2} & 25.3--32.7 & 10.5--10.6 \\
One-Shot & 44.8--53.3 & 18.1--25.2 & 8.30--12.0 & 8.20--12.4 &
\textbf{1.39--1.59} \\
\textbf{Agent} & 70.5--77.4 & 54.7--61.5 & 23.4--28.4 & 21.8--27.3 &
4.95--5.30 \\
\midrule
Oracle-Video & 88.1--92.3 & \textbf{70.5--77.6} & 29.9--34.9 &
\underline{27.1--33.1} & 4.81--5.13 \\
Oracle-Span & \textbf{98.2--99.5} & \underline{64.9--71.9} &
\textbf{45.2--51.2} & \textbf{44.7--51.3} & \underline{1.88--2.04} \\
\bottomrule
\end{tabular*}}

\end{table*}

\begin{table*}[t]
\centering
\caption{\textsc{DCI-QA} evidence-access diagnostics (GPT-5.5), using the
same frozen cases over three repetitions (345 ITT slots/row).  Entries
report only the endpoints of 52-parent-seed bootstrap 95\% confidence
intervals, each to three significant figures.  Ans.-5 averages available valid
GPT-5.4/GPT-5.6 item panels.  Bold and
underlined intervals follow the best and second-best underlying point
estimates, respectively, across rows.}
\label{tab:dci-qa-evidence-access-ci}
{\small
\setlength{\tabcolsep}{3pt}
\begin{tabular*}{\textwidth}{@{\extracolsep{\fill}}lcccccc@{}}
\toprule
Policy & \shortstack{SpanHit\\Rec. $\uparrow$} &
\shortstack{CiteV\\Rec. $\uparrow$} &
\shortstack{Span\\Cov. $\uparrow$} &
\shortstack{Span\\@.5 $\uparrow$} &
\shortstack{Ans.\\5 $\uparrow$} & Rounds $\downarrow$ \\
\midrule
\textbf{Full} & 32.7--45.0 & 38.4--51.9 & 7.50--12.6 & 4.70--10.6 &
2.49--2.83 & 7.82--10.2 \\
\midrule
Oracle-Video & \underline{53.0--64.3} & \underline{73.5--82.2} &
\underline{16.0--21.8} & \underline{12.2--20.3} &
\underline{2.66--2.99} & \underline{4.75--6.26} \\
Oracle-Span & \textbf{83.1--91.2} & \textbf{75.4--84.2} &
\textbf{63.3--73.2} & \textbf{61.1--71.5} &
\textbf{2.84--3.15} & \textbf{3.16--3.79} \\
\bottomrule
\end{tabular*}}

\end{table*}

\end{document}